\documentclass{article} 

\usepackage{amsmath,amsfonts,bm}

\def\eqref#1{equation~\ref{#1}}

\def\1{\bm{1}}

\DeclareMathAlphabet{\mathsfit}{\encodingdefault}{\sfdefault}{m}{sl}
\SetMathAlphabet{\mathsfit}{bold}{\encodingdefault}{\sfdefault}{bx}{n}

\usepackage{hyperref}
\usepackage{url}
\usepackage[]{natbib}
\usepackage[toc,page]{appendix}
\usepackage{graphicx}
\usepackage{amsmath}
\usepackage{amssymb}
\usepackage{hyperref}

\usepackage{textcomp}

\usepackage{pgfplots}
\usepackage{xcolor}
\usepackage{colortbl}

\usepackage{wrapfig}

\usepackage[accepted]{icml2019}

\definecolor{darkgreen}{rgb}{0,0.6,0}
\definecolor{darkred}{rgb}{0.7,0.0,0}
\definecolor{darkblue}{rgb}{0,0.0,0.6}

\newcommand{\sref}[1]{\S\ref{#1}}

\newcommand{\lcom}[1]{}
\newcommand{\ncom}[1]{}
\newcommand{\jcom}[1]{}
\newcommand{\jncom}[1]{}
\newcommand{\todo}[1]{}
\newcommand{\outline}[1]{}

\newcommand{\pp}[1]{\left( #1 \right)}

\newcommand{\mc}{\mathcal}

\newcommand{\expect}[2]{\mathbb{E}_{#1}\left[ #2 \right]}

\icmltitlerunning{Understanding and correcting pathologies in the training of learned optimizers}

\begin{document}

\twocolumn[
\icmltitle{Understanding and correcting pathologies in the training of learned optimizers}



\icmlsetsymbol{equal}{*}

\begin{icmlauthorlist}
\icmlauthor{Luke Metz}{goo}
\icmlauthor{Niru Maheswaranathan}{goo}
\icmlauthor{Jeremy Nixon}{goo}
\icmlauthor{C. Daniel Freeman}{goo}
\icmlauthor{Jascha Sohl-Dickstein}{goo}
\end{icmlauthorlist}

\icmlaffiliation{goo}{Google Brain}

\icmlcorrespondingauthor{Luke Metz}{lmetz@google.com}

\icmlkeywords{todo}

\vskip 0.3in
]



\printAffiliationsAndNotice{}  

\begin{abstract}
Deep learning has shown that learned functions can dramatically outperform hand-designed functions on perceptual tasks. 
Analogously, this suggests that learned optimizers may similarly outperform current hand-designed optimizers, especially for specific problems.
However, learned optimizers are notoriously difficult to train and have yet to demonstrate wall-clock speedups over hand-designed optimizers, and thus are rarely used in practice.
Typically, learned optimizers are trained by truncated backpropagation through an unrolled optimization process resulting in gradients that are either strongly biased (for short truncations) or have exploding norm (for long truncations).
In this work we propose a training scheme which overcomes both of these difficulties, by dynamically weighting two unbiased gradient estimators for a variational loss on optimizer performance, allowing us to train neural networks to perform optimization of a specific task faster than tuned first-order methods.
We demonstrate these results on problems where our learned optimizer trains convolutional networks faster in wall-clock time compared to tuned first-order methods and with an improvement in test loss.
\end{abstract}

\section{Introduction}
Gradient based optimization is a cornerstone of modern machine learning.
A large body of research has been targeted at developing improved gradient based optimizers.
In practice, this typically involves analysis and development of hand-designed optimization algorithms \citep{nesterov1983method,duchi2011adaptive,tieleman2012lecture,kingma2014adam}.
These algorithms generally work well on a wide variety of tasks, and are tuned to specific problems via hyperparameter search.
On the other hand, a complementary approach is to \textit{learn} the optimization algorithm \citep{bengio1990learning, schmidhuber1995learning, hochreiter2001learning, andrychowicz2016learning, wichrowska2017learned, li2017learning, lv2017learning, Bello17}.
That is, to learn a function that performs optimization, targeted at particular problems of interest.
In this way, the algorithm may learn \emph{task specific} structure, enabling dramatic performance improvements over more general optimizers.

However, training learned optimizers is notoriously difficult.
Existing work in this vein can be classified into two broad categories.
On one hand are black-box methods such as evolutionary algorithms~\citep{goldberg1988genetic, bengio1992optimization}, random search~\citep{bergstra2012random}, reinforcement learning~\citep{Bello17, li2016learning, li2017learning}, or Bayesian optimization~\citep{snoek2012practical}.
However, these methods scale poorly with the number of optimizer parameters.

\begin{figure*}[]
    \centering
    \includegraphics[width=5.5in]{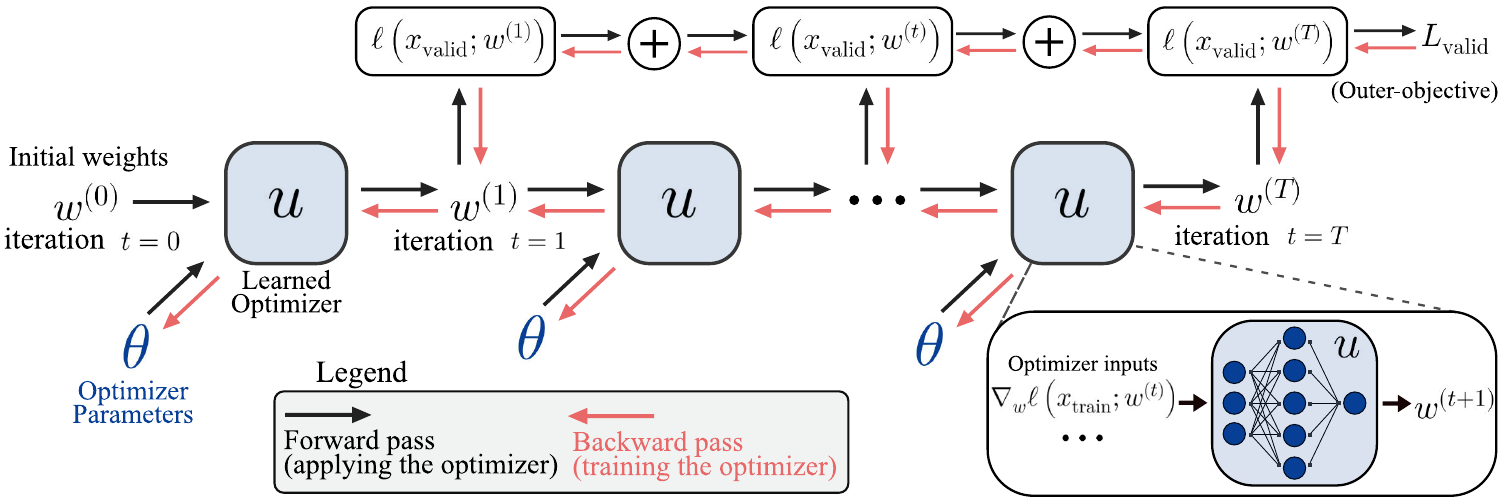}\\
    \vspace{4pt}
    \begin{tabular}{r|p{12cm}}
        \textbf{Term} & \textbf{Definition} \\
        &\\[-0.2cm]
        \hline
        &\\[-0.2cm]
        $\mathcal{D}$ & Dataset consisting of train and validation split, $\mc D_\text{train}$ and $\mc D_\text{valid}$. \\
        $\mc T$ & The set of tasks, where each task is a dataset (e.g., a subset of Imagenet classes).\\
        $w^{\pp{t}}$ & Parameters of inner-problem at iteration $t$. These are updated by the learned optimizer, and depend implicitly on $\theta$ and $\mc D_{\text{train}}$. 
        \\
        $\ell\pp{x; w^{\pp{t}}}$ & Loss on inner-problem, for mini-batch $x$. \\
        $\theta$ & Parameters of the optimizer. \\
        $u\pp{\cdot; \theta}
        $ & Function defining the learned optimizer. The inner-loop update is $w^{\pp{t+1}} = u\pp{w^{\pp{t}}, x, \nabla_{w} \ell, \ldots; \theta}$, for $x \sim {\mathcal{D}_\text{train}}$.\\
        $L_\text{train}\pp{\theta}$ & Outer-level objective targeting training loss, $\mathbb{E}_{\mathcal{D} \sim \mc T} \expect{x \sim \mathcal{D}_\text{train}}{\frac{1}{T} \sum_{t=1}^{T} \ell\pp{x; w^{\pp{t}}}}$.\\
        $L_\text{valid}\pp{\theta}$ & Outer-level objective targeting validation loss, $\mathbb{E}_{\mathcal{D} \sim \mc T} \expect{x \sim \mathcal{D}_\text{valid}}{\frac{1}{T} \sum_{t=1}^{T} \ell\pp{x; w^{\pp{t}}}}$.\\
        $\mc L\pp{\theta}$ & The variational (smoothed) outer-loop objective, 
        $\expect{\tilde{\theta} \sim \mc N\pp{\theta, \sigma^2 I}}{L\pp{\tilde{\theta}}}$.
    \end{tabular}
    \caption{\textbf{Top:} Schematic of unrolled optimization. \textbf{Bottom:} Definition of terms used in this paper.}
    \label{tab:my_label}
\end{figure*}

The other approach is to use first-order methods, by computing the gradient of some measure of optimizer effectiveness with respect to the optimizer parameters.
Computing these gradients is costly as we need to iteratively apply the learned update rule, and then backpropagate through these applications, a technique commonly referred to as ``unrolled optimization''~\citep{bengio2000gradient, maclaurin2015gradient}. 
To address the problem of backpropagation through many optimization steps (analogous to many timesteps in recurrent neural networks), many works make use of truncated backpropagation though time (TBPTT) to partition the long unrolled computational graph into separate pieces~\citep{werbos1990backpropagation, domke2012generic, tallec2017unbiasing}.
This not only yields computational savings, at the cost of increased bias~\citep{tallec2017unbiasing}, but also limits exploding gradients which emerge from too many iterated non-linear function applications~\citep{pascanu2013difficulty, parmas2018pipps}.
Existing methods have been proposed to address the bias of TBPTT but come at the cost of increased variance or computational complexity \citep{williams1989learning, ollivier2015training, tallec2017unbiasing}.
Previous techniques for training RNNs via TBPTT have thus far not been effective for training optimizers.

In this paper, we analytically and experimentally explore the debilitating role of bias and exploding gradients on training optimizers (\sref{sec bias explosion}). 
We then show how these pathologies can be remedied by optimizing the parameters of a distribution over the optimizer parameters, known as variational optimization~\citep{staines2012variational} (\sref{sec more stable}).
We define two unbiased gradient estimators for this objective: a reparameterization based gradient \citep{kingma2013auto}, and evolutionary strategies \citep{rechenberg1973evolutionsstrategie, nesterov2011random}.
By dynamically reweighting the contribution of these two gradient estimators \citep{fleiss1993review, parmas2018pipps, buckman2018sample}, we are able to avoid exploding gradients and unroll longer, to stably and efficiently train learned optimizers.

We demonstrate the utility of this approach by training a learned optimizer to target optimization of small convolutional networks on image classification (\sref{sec:experiments}). With our method, we are able to outer-train on more inner steps (10k inner-parameter updates) with more complex inner-problems than prior work.
Additionally, we can simplify the parametric form of the optimizer, utilizing a small MLP without any complex tricks such as extensive use of normalization\citep{metz2018learning}, or annealing training from existing algorithms\citep{houthooft2018evolved} previously needed for stability.

On the targeted task distribution, this learned optimizer achieves better test loss, and is faster in \emph{wall-clock time}, compared to hand-designed optimizers such as SGD+Momentum, RMSProp, and ADAM (Figure \ref{fig:unroll_on_problem}).
To our knowledge, this is the first instance of a learned optimizer performing comparably to existing methods on wall-clock time, as well as the first parametric optimizer outer-trained against validation loss.
While not the 
focus of this work, we also find that the learned optimizer demonstrates promising generalization ability on out of distribution tasks (Figure \ref{fig:outofdomain}).

\section{Unrolled optimization for learning optimizers}
\subsection{Problem Framework}
Our goal is to learn an optimizer which is well suited to some set of target optimization tasks. 
Throughout the paper, we will use the notation defined in Figure \ref{tab:my_label}.
Learning an optimizer can be thought of as a bi-level optimization problem \citep{franceschi2018bilevel}, with {\em inner} and {\em outer} levels.
The inner minimization consists of optimizing the weights ($w$) of a target problem $\ell(w)$ by the repeated application of an update rule ($u\pp{\cdot}$).
The update rule is a parameterized function that defines how to map the weights at iteration $t$ to iteration $t+1$: $w^{(t+1)} = u(w^{(t)}, x, \nabla_{w} \ell, ...; \theta)$.
Here, $\theta$ represents the parameters of the learned optimizer.
In the outer loop, these optimizer parameters ($\theta$) are updated so as to minimize some measure of optimizer performance, the outer-objective ($L(\theta)$). 
Our choice for
$L$ will be the average value of the target loss ($\ell\pp{\cdot}$) measured over either training or validation data.
Throughout the paper, we use \textit{inner-} and \textit{outer-} prefixes to make it clear when we are referring to applying a learned optimizer on a target problem (inner) versus training a learned optimizer (outer).

\begin{figure*}
    \centering
    \includegraphics[width=6.65in]{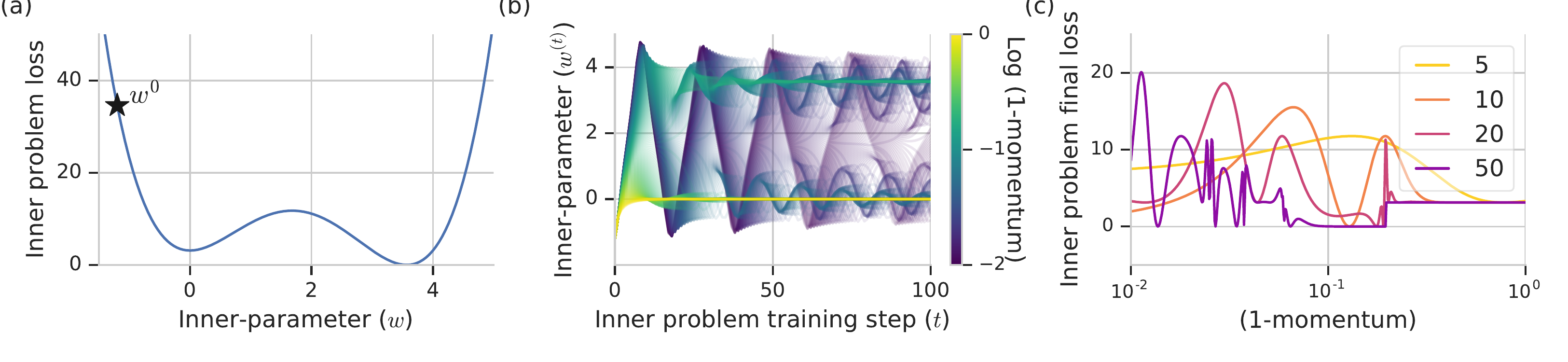}\vspace{-5pt}
    \includegraphics[width=6.65in]{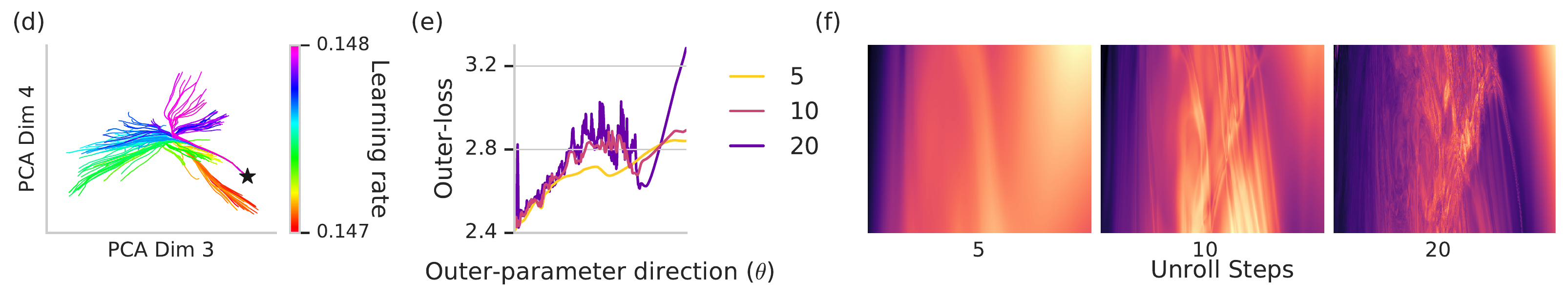}

    \caption{
    Outer-problem optimization landscapes can become increasingly pathological with increasing inner-problem step count. 
    \textbf{(a)} A toy 1D inner-problem loss surface with two local minimum. Initial parameter value ($w^{(0)}$) is indicated by the star. \textbf{(b)} Final inner-parameter value ($w^{(T)}$) as a function of the number of inner problem training steps $T$, when inner-problem training is performed by SGD+momentum. Color denotes different values of the momentum parameter. Low momentum (yellow) converge to the first local minimum at $w=0$. Slightly higher momentum (green) escape this minimum to settle at the global minimum ($w\approx3.5$).
    Even larger values (purple) oscillate before eventually settling in one of the two minima.
    \textbf{(c)} The final loss after some number steps of optimization (shown in different colors) as a function of the momentum.
    The final loss surface is smooth for small number of training steps $T$. 
    However, larger values of $T$ result in near discontinuous loss surfaces around the transition points between the two minima.
    \textbf{(d)} Random 2D projection of the parameters of a small MLP trained with Adam using different learning rates. As more steps are taken, the inner-training trajectories diverge.
    \textbf{(e)} 
    Similar to \textbf{c}, where the inner problem is a two layer MLP, the learned optimizer is the one used in this paper, and for a 1D slice through the outer-parameters $\theta$ along the gradient direction. The outer-objective shown is average validation loss for 5, 10, and 20 inner-unrolls.
    \textbf{(f)} 
    2D rather than 1D slices through $\theta$, for different numbers of inner-loop steps. Intensity indicates value of $L_\text{train}\pp{\theta}$; darker is lower.
    Similar pathologies are observed to those which manifest in the toy problem.
    \label{fig:real_loss_surface}
    \label{fig:2minimum}
    }
\end{figure*}

\subsection{Unrolled optimization}
In order to train an optimizer, we wish to compute derivatives of the outer-objective $L$ with respect to the optimizer parameters, $\theta$.
Doing this requires unrolling the optimization process.
That is, we can form an unrolled computational graph that consists of iteratively applying an optimizer ($u$) to optimize the weights ($w$) of a target problem (Figure~\ref{tab:my_label}).
Computing gradients for the optimizer parameters involves backpropagating the outer loss through this unrolled computational graph.
This is a costly operation, as the entire inner-optimization problem must be unrolled in order to get a single outer-gradient.
Partitioning the unrolled computation into separate segments, known as truncated backpropagation, allows one to compute multiple outer-gradients over shorter segments.
That is, rather than compute the full gradient from iteration $t=0$ to $t=T$, we compute gradients in windows from $t=a$ to $t=a+\tau$.
The gradients from these segments can be used to update $\theta$ \emph{without} unrolling all $T$ iterations, dramatically decreasing the computation needed for each update to $\theta$.
The choice for the number of inner-steps per truncation is challenging. Using a large number of steps per truncation can result in exploding gradients making outer-training difficult, while using a small number of steps can produce biased gradients resulting in poor performance. In the following sections we analyze these two problems.

\subsection{Exponential explosion of gradients with increased sequence length}\label{sec bias explosion}
We can illustrate the problem of exploding gradients analytically with a simple example: learning a learning rate.
Following the notation in Figure~\ref{tab:my_label}, we define the optimizer as:
$$ w^{(t+1)} = u(w^{(t)}; \theta) = w^{(t)} - \theta \nabla \ell\pp{w^{(t)}}, $$
where $\theta$ is a scalar learning rate that we wish to learn for minimizing some target problem $\ell(w^{(t)})$. For simplicity, we assume a deterministic loss ($\ell(\cdot)$) with no batch of data ($x$).

The quantity we are interested in is the derivative of the loss after $T$
steps of gradient descent with respect to $\theta$.
We can compute this gradient (see Appendix \ref{app:derivation}) as:
$$ \frac{d \ell(w^{(T)})}{d\theta} = \left\langle g^{(T)}, -\sum_{i = 0}^{T-1} \left(\prod_{j=i+1}^{T-1} (I - \theta H^{(j)}) \right) g^{(i)}\right\rangle, $$
where $g^{(i)}$ and $H^{(j)}$ are the gradient and Hessian of the target problem $\ell(w)$ at iteration $i$ and $j$, respectively.
We see that this equation involves a sum of products of Hessians.
In particular, the first term in the sum involves a product over the \textit{entire sequence} of Hessians observed during training.
When optimizing a quadratic loss, the Hessian is constant, and the outer-gradient becomes a matrix polynomial of degree $T$, where $T$ is the number of gradient descent steps.
Thus, the outer-gradient can grow exponentially with $T$ if the maximum eigenvalue of $(I-\theta H^{(j)})$ is greater than $1$.
In general, the Hessian is not 
a constant, 
but in practice we still see an exponential growth in gradient norm across a variety of settings.

We can see another problem with long unrolled gradients empirically.
Consider the task of optimizing a loss surface with two local minima defined as $\ell(w) = (w-4)(w-3)w^2$ with initial condition $w^{(0)}=-1.2$ using a momentum based optimizer with a parameterized momentum value $\theta$ (Figure \ref{fig:2minimum}a).
At low momentum values the optimizer converges in the first of the two local minima, whereas for larger momentum values the optimizer settles in the second minimum.
With even larger values of momentum, the iterate oscillates between the two minima before settling.
We visualize both the trajectory of $w^{\pp{t}}$ over training and the final loss value for different momentum values in Figure \ref{fig:2minimum}b and \ref{fig:2minimum}c.
With increasing unrolling steps, the loss surface as a function of the momentum $\theta$ becomes less and less smooth, and develops near-discontinuities at some values of the momentum resulting in extremely large gradient norms. This behavior is not unique to momentum optimizer parameters. In Appendix \ref{app:toy_lr} we perform additional experiments modifying learning rates instead and show similar behavior.

Although these are toy systems,
we see similar pathologies 
when optimizing more complex inner-models with both hand designed and learned optimizers.
Trajectories taken during optimization often change dramatically as a result of only small changes in optimizer parameters. 
To illustrate this, in Figure \ref{fig:2minimum}d we train a 2 layer MLP with ReLU activations for 40 iterations using Adam. We vary the learning rate between 0.1469 to 0.1484 with 100 samples spaced uniformly in log scale, but keep all sources of randomness fixed. We plot 2D random projections of the MLP's parameters during training, using color to denote different learning rates. Early in training the trajectories are similar, but as more steps are taken the trajectories diverge, producing drastically different trained models with only small changes in the learning rate.

In the case of both neural network inner-problems and neural network optimizers, the outer-loss surface can grow even more complex with increasing number of unrolling steps. We illustrate this in Figure \ref{fig:real_loss_surface}e and \ref{fig:real_loss_surface}f for slices through the loss landscape $L\pp{\theta}$ of the outer-problem for a neural network optimizer.

\begin{figure*}[t]
    \centering
    \includegraphics[width=6.5in]{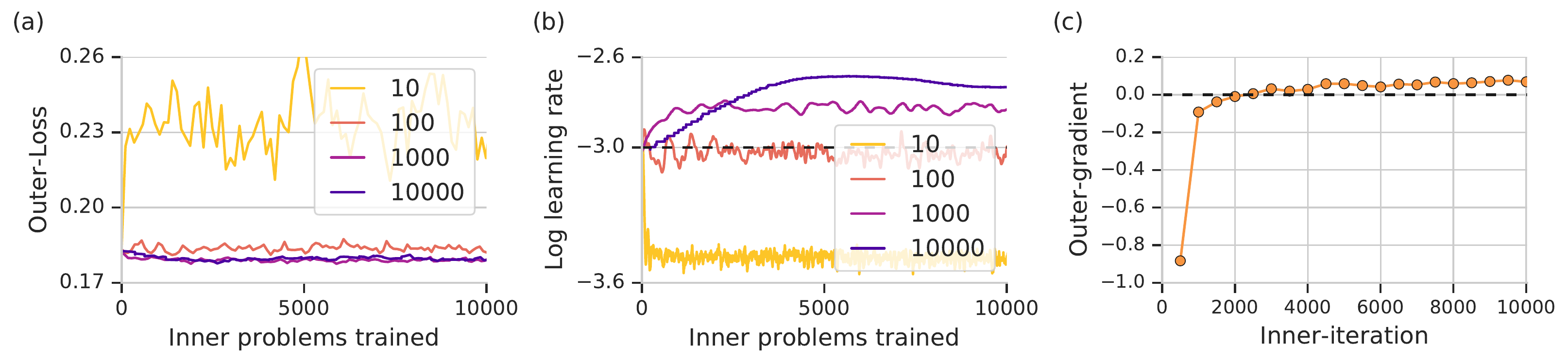}        
    \caption{
    Large biases can result from reducing the number of steps per truncation in unrolled optimization.
     \textbf{(a)} Each line represents an experiment with the same total number of inner-steps (10k) but a different number of unrolling steps per truncation.
     For each truncation amount, we show an exponential rolling average of outer-loss obtained while outer-training. Short truncations converge to suboptimal outer-loss, while longer truncations (1000, 10000) are less biased and thus achieve lower outer-loss.
    \textbf{(b)} In all cases, the initial learning rate is 0.001 (dashed line).
    We find low truncation amounts move away from the optimal learning rate, increasing the outer-loss, while larger truncation amounts increase the learning rate, decreasing the outer-loss.
    \textbf{(c)} Outer-gradient at each truncation over the course of inner-training.
    Initial outer-gradients are highly negative, trying to increase the learning rate, while later outer-gradients are slightly above zero, decreasing learning rate.
    The total outer-gradient is the sum of these competing contributions.
    \label{fig:meta_train_learning_rate}
    \label{fig:lr_gradient_vis}
    }
\end{figure*}

\subsection{Increasing bias with truncated gradients} \label{sec:increase_bias}
Existing work on learned optimizers often avoids exploding gradients (\sref{sec bias explosion}) by using a short truncation window. Here, we demonstrate the bias short truncation windows can introduce in unrolled optimization. 
These results are similar to those presented in \citet{wuunderstanding}, except that we utilize \emph{multiple} truncations rather than a single, shortened unroll.
First, consider outer-learning the learning rate of Adam when optimizing a small two layer neural network on MNIST \citep{lecun1998mnist}.
We initialize Adam with a learning rate of $0.001$ and outer-train using increasing truncation amounts (Figure \ref{fig:meta_train_learning_rate}ab). Adam is used as the outer-optimizer. Other outer-optimizers can be found in \ref{app:trunc_bias}.
Despite initializing close to the optimal learning rate, when outer-training with severely truncated backprop the resulting learning rate decreases, increasing the outer-loss. The sum of truncated outer-gradients are anti-correlated with the true outer-gradient.

\section{Towards stable training of learned optimizers}

\label{sec more stable}
To perform outer-optimization of a loss landscape with high frequency structure like that in Figure~\ref{fig:real_loss_surface}, one might intuitively want to smooth the outer-objective loss surface.
To do this, instead of optimizing $L(\theta)$ directly we instead optimize a smoothed outer-loss $\mc L\pp{\theta}$,
\begin{equation*}
\mc L\pp{\theta} = \expect{\tilde{\theta} \sim \mc N\pp{\theta, \sigma^2 I}}{L\pp{\tilde{\theta}}},
\end{equation*}
where $\sigma^2$ is a fixed variance (set to 0.01 in all experiments) which determines the degree of smoothing. This is the same approach taken in variational optimization~\citep{staines2012variational}.
We can construct two different unbiased gradient estimators for $\mc L\pp{\theta}$: one via the reparameterization trick \citep{kingma2013auto}; and one via the ``log-derivative trick'', similarly to what is done in evolutionary strategies (ES) and REINFORCE \citep{williams1992simple, wierstra2008natural}.
We denote the two estimates as $g_\text{rp}$ and $g_\text{es}$ respectively,
\begin{alignat*}{4}
    g_\text{rp} &= 
    \frac{1}{S}\sum_s {\nabla_{\theta} L\pp{\theta+\sigma n_s}},\\
    \qquad & n_s \sim N\pp{0, I},&
    \\
    g_\text{es} &= 
    \frac{1}{S}\sum_s 
     L\pp{\tilde{\theta}_s} \nabla_{\theta} \left[\text{log}\pp{N\pp{\tilde{\theta}_s; \theta, \sigma^2I}}\right],\\
     \qquad & \tilde{\theta}_s \sim N\pp{\theta, \sigma^2 I},&
\end{alignat*}

where $N\pp{\tilde{\theta}_s; \theta, \sigma^2I}$ is the probability density of the given ES sample, $\tilde{\theta}_s$, $S$ is the sample count, and in implementation the same samples can be reused for $g_{rp}$ and $g_{es}$.

\begin{figure}[t]
    \centering
    \includegraphics[width=3.2in]{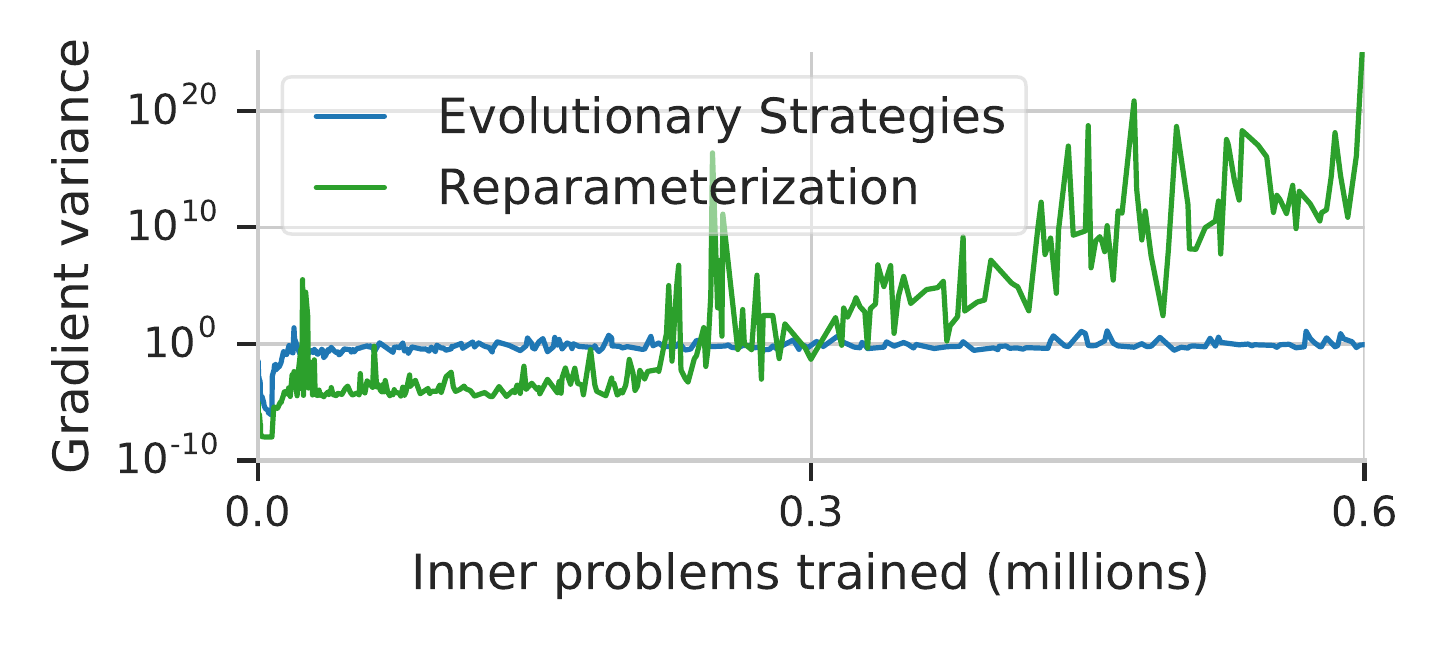}
    \vspace{-10pt}
    \caption{As outer-training progresses, the variance of the reparameterization gradient estimator grows, while the variance of the ES estimator remains constant. The variance of the reparameterization gradient progresses from approximately five orders of magnitude better than that of the ES gradient, to approximately twenty orders of magnitude worse.
        \label{fig:grad_var}
        }
\end{figure}

Following the insight from \citep{parmas2018pipps} in the context of reinforcement learning\footnote{
  \cite{parmas2018pipps} go on to propose a more sophisticated gradient estimator that operates on a per iteration level.
  While this should result in an even lower variance estimator in our setting, we find that the simpler solution of combing both terms at the end is easier to implement and works well in practice.},
we combine these estimates using inverse variance weighting \citep{fleiss1993review},
\begin{equation}\label{eq:combine}
    g_\text{merged} = \frac{g_\text{rp}\sigma_\text{rp}^{-2} + g_\text{es}\sigma_\text{es}^{-2}}{\sigma_\text{rp}^{-2}+\sigma_\text{es}^{-2}},
\end{equation}
where $\sigma_\text{rp}^2$ and $\sigma_\text{es}^2$ are empirical estimates of the variances of $g_{rp}$ and $g_{es}$ respectively.
When outer-training learned optimizers we find the variances of $g_\text{es}$ and $g_\text{rp}$ can differ by as many as 20 orders of magnitude (Figure~\ref{fig:grad_var}).
This merged estimator addresses this by having at most the lowest of the two variances. 
To further reduce variance, we employ antithetic sampling. Each normal distribution draw is used twice, both positive and negative, when computing $g_\text{rp}$ and $g_\text{es}$.

The cost of computing a single 
sample
of $g_\text{es}$ and $g_\text{rp}$ is thus two forward and two backward passes of an unrolled optimization.
To compute the empirical variance, we leverage data parallelism to compute multiple samples of $g_\text{es}$ and  $g_\text{rp}$.
In theory, to prevent bias the samples used to evaluate $\sigma_\text{rp}^2$ and $\sigma_\text{es}^2$ must be independent of those used to estimate $g_\text{es}$ and $g_\text{rp}$, but in practice we found good performance using the same samples for both.

This gradient estimator fixes the exploding gradients problem when computing gradients over a long truncation, and longer truncations enable lower bias gradient estimates.
In practice, 
these longer truncations are computationally expensive,
and early in outer-training shorter truncations
are sufficient.
The full outer-training algorithm is described in Appendix \ref{app:algo}.

\section{Experiments} \label{sec:experiments}
As a proof of principle, we use the training algorithm described in \sref{sec more stable} to train a simple learned optimizer. 
For this work, we focus on training an optimizer to target a specific architecture. 
In the following sections we describe the optimizer architecture used, the task distribution 
on which 
we outer-train, as well as outer training details. 
We then discuss the performance of our 
learned optimizer on
both
in and out of distribution target problems. 
We finish with an ablation study showing the importance of our gradient estimator, as well as aspects of the optimizer's architecture.

\begin{figure}
    \centering
    \includegraphics[width=3.4 in]{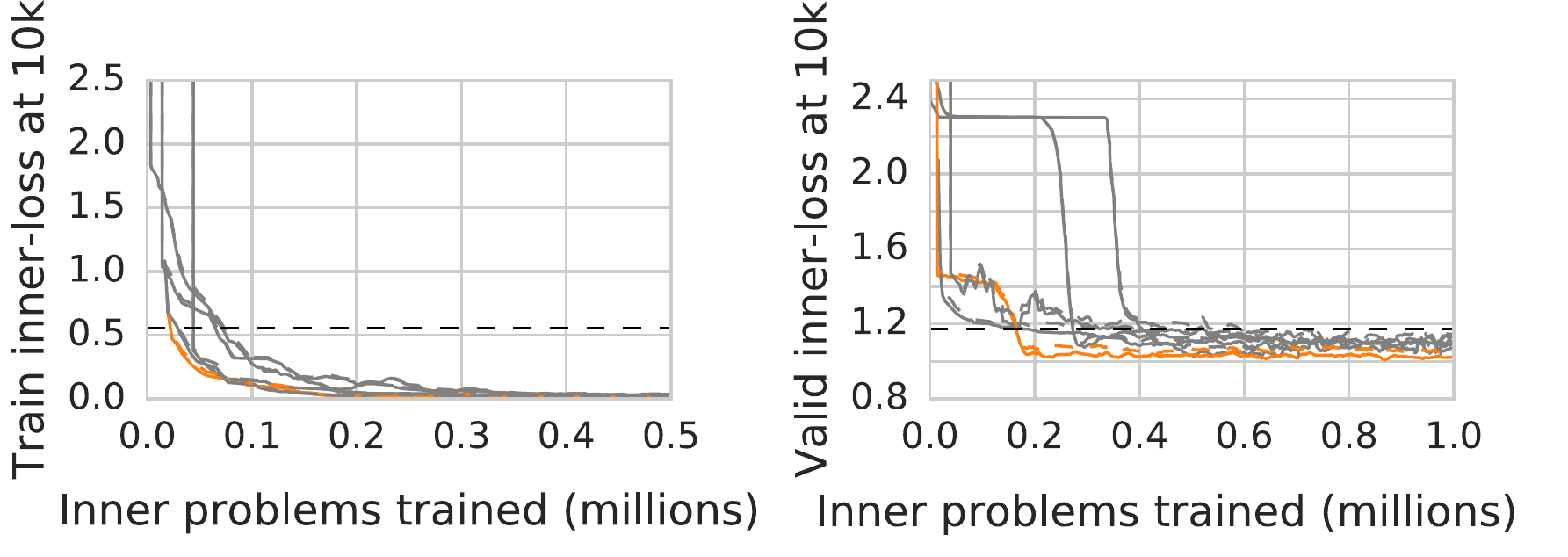}
    \vspace{-10pt}
    \caption{
    Performance after 10k iterations of inner-training for different 
    learned optimizers 
    over the course of outer-training. Each line represents a different random initialization of outer-parameters ($\theta$). Optimizers are trained targeting the train outer-objective (left), and the validation outer-objective (right). Dashed lines indicate performance of learning rate tuned Adam.
    Models in orange are the best performing and used in \sref{sec:experiments}.
    \label{fig:meta-training-curves}
    }
\end{figure}

\subsection{Optimizer architecture} \label{sec:architecture}
The optimizer architecture used in all experiments consists of a small, fast to compute, fully connected neural network, with one hidden layer containing 32 ReLU units ($\sim$1k parameters). This network is applied to each target problem inner-parameter independently.
The outputs of the MLP consist of an un-normalized update direction and a per parameter log learning rate which gets exponentiated. These two quantities are multiplied and subtracted from the previous inner-parameter value to form the next inner-parameter value.
The MLP for each weight takes as input: the gradient with respect to that weight, the parameter value, 
exponentially weighted moving averages of gradients at multiple time scales \citep{lucas2018aggregated}, as well as a representation of the current iteration number.
Many of these input features were motivated by \cite{wichrowska2017learned}.
We conduct ablation studies for these inputs in \sref{sec:ablations}.
See Appendix \ref{app:arch_details} for further architectural details.

\begin{figure*}[t]
    \centering
    \includegraphics[width=6.8in]{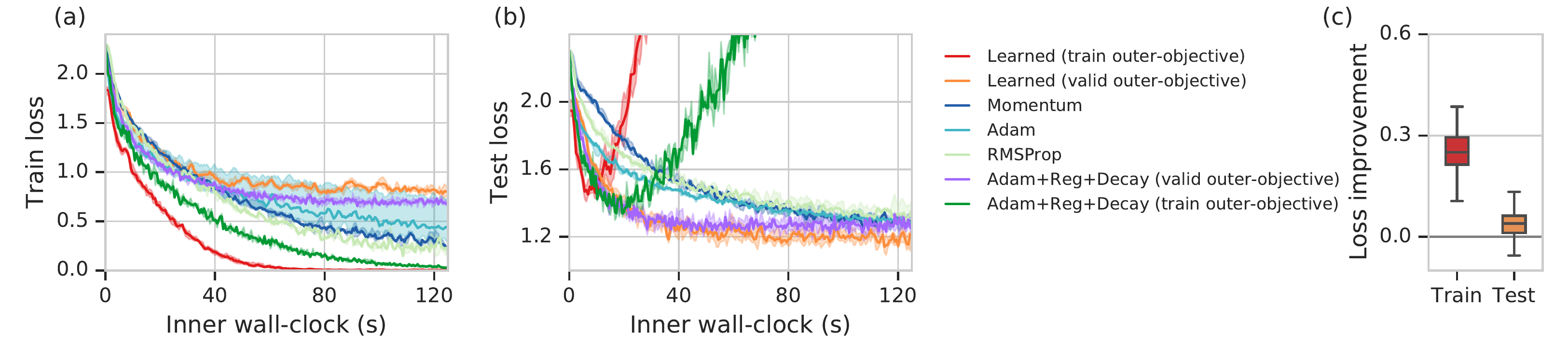}

    \caption{
    Learned optimizers outperform existing optimizers on training loss (a) and test loss (b).
    \textbf{(a,b)}
    Training and test curves on outer-test tasks (tasks not seen during outer-training).
    We show two learned optimizers -- one trained to minimize training loss, and the other trained to minimize validation loss on the inner-problem. 
    We provide 2 types of baselines. First, learning rate tuned Adam, RMSProp, and SGD+Momentum, for both train and validation loss (Panel (a) and (b), respectively).
    Second, we show tuned Adam (learning rate, beta1, beta2, epsilon), with learning rate schedules (linear decay and exponential decay values), and regularization (multipliers on l1 and l2 added to the inner-loss). We tune with random search against both train and validation loss.
    On training loss (a), our learned optimizer approaches zero training loss faster than existing methods.
    On test loss (b), our learned optimizer is of similar speed, but converges to a 
lower minimum.
    Shaded regions correspond to 25 and 75 percentile over five random initializations of the CNN.
    For plots showing performance in terms of step count rather than wall-clock, and for more task instances, see Appendix \ref{app:inner_loop_figure}. See Appendix \ref{app:wallclock} for details on the wall-clock calculations.
    \textbf{(c)} 
    Distribution of the performance difference between the learned optimizers and
    corresponding Adam+Reg+Decay baseline. Positive values indicate performance better than baseline. We show training and test losses for the outer-testing task distribution (tasks not seen during training of the optimizer).
    On the majority of tasks, the learned optimizers outperform the baseline.
    \label{fig:unroll_on_problem}
    \label{fig:histogram}
    }
    \vspace{-3pt}
\end{figure*}

\subsection{Optimizer target problem}\label{sec:target_problem}
The problem that each learned optimizer is trained against ($\ell\pp{\cdot}$) consists of training a three layer convolutional neural network (32 units per layer, ~20k parameters) inner-trained for ten thousand inner-iterations on 32x32x3 image classification tasks. Due to the weight sharing in convolutions, the compute per parameter is high and thus relatively less computation is needed for each inner-iteration.
We split the Imagenet dataset \citep{ILSVRC15} by class into 700 training and 300 test classes, and sample training and validation problems by sampling 10 classes at random using all images from each class.
This experimental design lets the optimizer learn problem specific structure (e.g. convolutional networks trained on object classification), but does not allow the optimizer to memorize class-specific weights for the base problem.
This task is modeled after the fact that standard architectures, e.g. ResNet \citep{he2016identity}, are often applied to a variety of different datasets.
See Appendix \ref{app:arch_details} for further details.

\subsection{Outer-training} \label{sec:curriculum}
To train the optimizer, we linearly increase the number of unrolled steps from 50 to 10,000 over the course of 5,000 outer-training weight updates.
The number of unrolled steps is additionally jittered by a small percentage (sampled uniformly up to 20\%).\
Due to the heterogeneous, small iterated computations, we train with asynchronous, batched SGD using 128 CPU workers.

Figure~\ref{fig:meta-training-curves} shows the performance of the optimizer (averaged over 40 randomly sampled outer-train and outer-test inner-problems) while outer-training.
Despite the stability improvements described in the last section, there is still variability in optimizer performance over random initializations of the optimizer parameters.
As expected given our optimizer parameterization, there is very little outer-overfitting. Nevertheless, we use outer-training loss to select the best model and use this in the remainder of the evaluation.

\subsection{Learned optimizer performance} \label{sec:learned_opt_performance}
Figure~\ref{fig:unroll_on_problem} shows performance of the learned optimizer, after outer-training, compared against other first-order methods on a sampled validation task (classes not seen during outer-training).
For ``Adam'', ``RMSProp'', and ``Momentum'', we report the best performance after tuning the learning rate by grid search using 11 values over a logarithmically spaced range from $10^{-4}$ to 10 on a per task basis. 
Searching over fixed learning rates is 
the baseline most commonly used 
in other learned optimizer work \citep{andrychowicz2016learning, wichrowska2017learned}. 
However, practitioners often adjust many more hyperparameters. 
Therefore, we provide an additional baseline consisting of tuning: 
all Adam optimizer parameters (beta1, beta2, epsilon, learning rate);
learning rate decay (exponential decay coefficient, linear decay coefficient); and regularization parameters (l1 regularization, l2 regularization). 
We outer-optimize these parameters against the distribution of tasks, evaluating \textgreater 2k parameter combinations totalling \texttildelow 100k total inner-problem training runs (performance is averaged over 80 inner runs for a lower variance estimate of $L$).
As with the learned optimizer, we outer-optimize with respect to $L_\text{train}$, and $L_\text{test}$. 
We outer-optimize our baseline using uniform random search over all 8 hyperparameters. Outer-optimizing the baseline with the framework we present in this paper yields similar results both in compute cost, and final performance achieved. We performed a similar optimization for RMSProp and SGD+Momentum and find similar performance to that of Adam. Results are not shown here for clarity but can be found in Appendix \ref{app:inner_loop_figure}.

When outer-trained against the training outer-objective, $L_\text{train}$, our learned optimizer achieves faster convergence on training loss (Figure~\ref{fig:unroll_on_problem}a), but poor performance on test loss (Figure~\ref{fig:unroll_on_problem}b). This is expected, as our outer-training procedure never sees validation loss, and thus only minimizes training loss, causing overfiting.
When outer-trained against the validation outer-objective, $L_\text{valid}$, we also achieve fast optimization and reach a lower test loss in the given time interval (Figure~\ref{fig:unroll_on_problem}b).
We suspect further gains in validation performance could be obtained with the addition of more regularization techniques into both the learned optimizer, and the tuned Adam baseline.

Figure~\ref{fig:histogram}c summarizes the performance of the learned optimizer across 100 sampled outer-test tasks (tasks not seen during outer-training).
It shows the difference in loss (averaged over the first 10k iterations of training) between the learned optimizer and the 8 parameter Adam (Adam+Reg+Decay) baseline tuned against the corresponding loss.
Our learned optimizer outperforms this baseline on the majority of tasks.

\begin{figure*}
    \centering
    \includegraphics[width=5.5in]{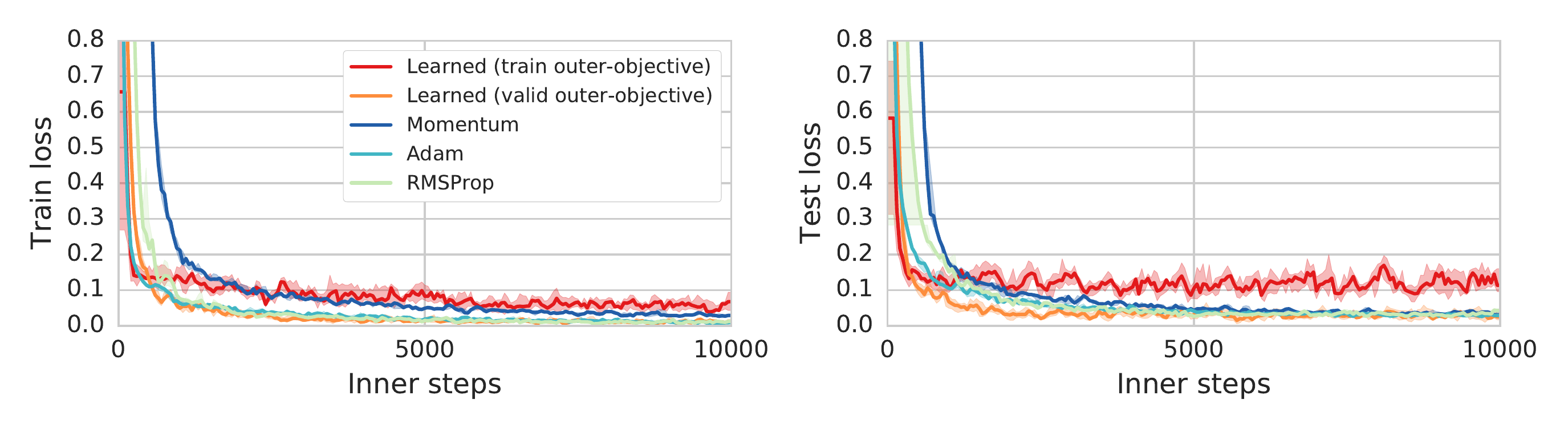}
    \vspace{-10pt}
    \caption{
    Train and test learning curves for an out of distribution training task consisting of a six layer convolutional neural network trained on 28x28 MNIST. We compare against learning rate tuned Adam, RMSProp, and SGD+Momentum.
    Note that the optimizer trained to target validation loss generalizes better than the one trained to target train loss. See Appendix \ref{app:outofdist} for experiments testing generalization to additional tasks.
    \label{fig:outofdomain}
    }
\end{figure*}

Although the focus of our approach was \emph{not} generalization, we find that our learned optimizer nonetheless generalizes to varying degrees to 
dissimilar datasets, different numbers of units per layer, different number of layers, and even to fully connected networks. In Figure \ref{fig:outofdomain} we show performance on a six layer convolutional neural network trained on MNIST. Despite the different number of layers, dissimilar dataset, and different input size, the learned optimizers still reduces the loss, and in the case of the validation outer-objective trains faster and generalizes well. We further explore the limits of generalization of our learned optimizer on additional tasks in Appendix \ref{app:outofdist}.

\subsection{Ablations} \label{sec:ablations}
\begin{figure*}
    \centering
    \includegraphics[width=5.3in]{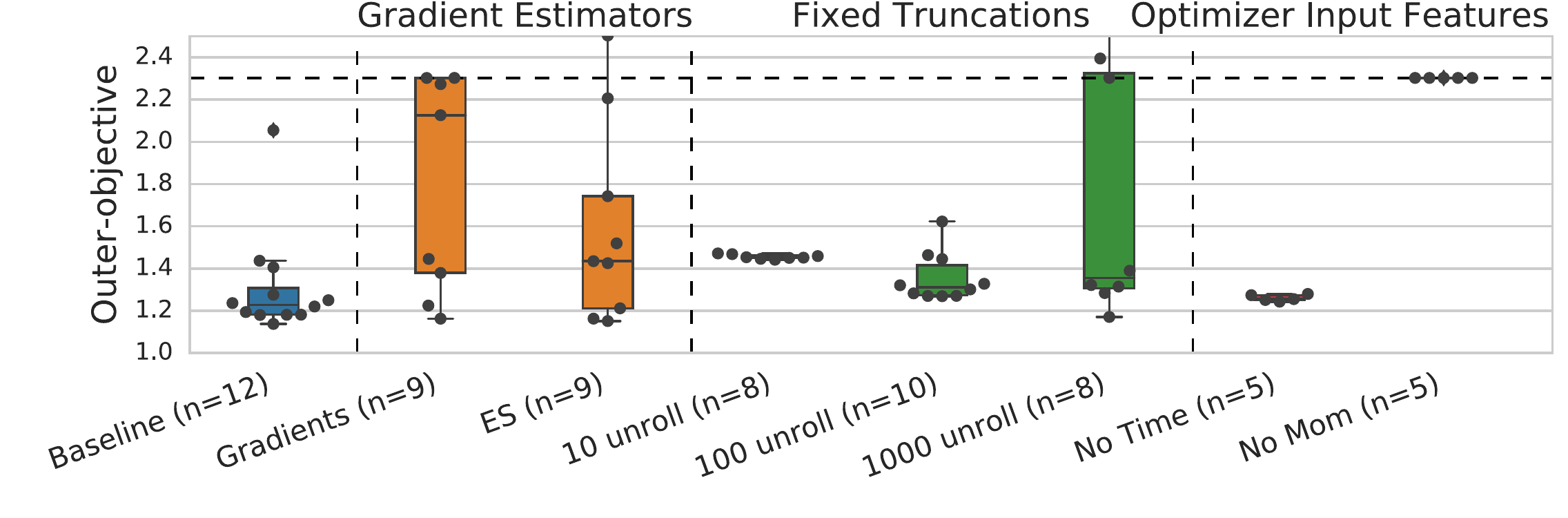}
    \vspace{-10pt}
    \caption{
    Ablation study. We compare the model described in \sref{sec:architecture} with different features removed. Shown above is the distribution of outer-loss performance across $n$ random seeds averaged between 380k and 420k inner problem trained for each seed.
    For full learning curves, see Appendix \ref{app:ablation}.
    We find the combined gradient estimator is more stable and outperforms both the analytic gradient, and evolutionary strategies. 
    To test the role of the truncation length curriculum, we show different, fixed truncation lengths. We find poor convergence with shorter unrolls (10, 100 steps per unroll) and high variance with longer unrolls (1000 steps per unroll).
    Finally, we find our learned optimizers perform 
    nearly as well
    \emph{without} the training step as an input,
    but fails to converge when not given access to momentum based features.
    \label{fig:ablations}
    }
\end{figure*}

To assess the importance of the gradient estimator discussed in \sref{sec more stable}, the unrolling curriculum \sref{sec:curriculum}, as well as the features fed to the optimizer enumerated in \sref{sec:architecture}, we re-trained the learned optimizer removing each of these additions.
In particular, we trained optimizers with: only the reparameterization gradient estimator (Gradients), only with evolutionary strategies (ES), a fixed number unrolled steps per truncation (10, 100, 1000) as opposed to a schedule keeping while keeping the same total inner-weight updates, no momentum terms (No Mom), and without the current iteration (No Time). To account for variance, each configuration is repeated with multiple random seeds.
Figure~\ref{fig:ablations} summarizes these findings, showing the learned optimizer performance for each of these ablations.
We find that the gradient estimator (in \sref{sec more stable}) and an increasing schedule of unroll steps are critical to performance, along with including momentum as an input to the optimizer.

\section{Discussion}
In this work we demonstrate two difficulties when training learned optimizers: ``exploding'' gradients, and a bias introduced by truncated backpropagation through time.
To combat this, we construct a variational bound of the outer-objective and minimize this via a combination of reparameterization and ES style gradient estimators.
By using our combined estimator and a curriculum over truncation step we are able to train learned optimizers that are faster in wall-clock time compared to existing optimizers.

In this work, we focused on applying optimizers to a restricted family of tasks.
While useful in its own right (e.g. rapid retraining of models on new data), future work will explore the limits of ``no free lunch''~ \citep{wolpert1997no} in the context of optimizers, to understand how and when learned optimizers generalize across tasks.
We are also interested in using these methods to better understand what problem structure our learned optimizers exploit.
By analyzing the trained optimizer, we hope to develop insights that may transfer back to hand-designed optimizers.
Outside of meta-learning, we believe the outer-gradient estimator presented here can be used to train other long time dependence recurrent problems such as neural turning machines~\citep{graves2014neural}, or neural GPUs~\citep{kaiser2015neural}. 

Much in the same way deep learning has replaced feature design for perceptual tasks, we see meta-learning as a tool capable of learning new and interesting algorithms, especially for domains with unexploited problem-specific structure.
With better outer-training stability, we hope to improve our ability to learn interesting algorithms, both for optimizers and beyond.

\subsubsection*{Acknowledgments}
We would like to thank
Madhu Advani,
Alex Alemi,
Samy Bengio,
Brian Cheung,
Chelsea Finn,
Sam Greydanus,
Hugo Larochelle,
Ben Poole,
George Tucker,
and 
Olga Wichrowska, 
as well as the rest of the Brain Team for conversations that helped shape this work.

\bibliography{main}
\bibliographystyle{icml2019}

\newpage

\normalsize
\onecolumn
\appendix

\section{Derivation of the unrolled gradient} \label{app:derivation}
For the case of learning a learning rate, we can derive the unrolled gradient to get some intuition for issues that arise with outer-training. Here, the update rule is given by:
$$ w \leftarrow w - \theta \nabla \ell(w), $$
where $w$ are the inner-parameters to train, $\ell$ is the inner-loss, and $\theta$ is a scalar learning rate, the only outer-parameter. We use superscripts to denote the iteration, so $w^{(t)}$ are the parameters at iteration $t$. In addition, we use $g^{(t)} = \nabla \ell(w^{(t)})$ and $H^{(t)} = \nabla^2 \ell(w^{(t)})$ to denote the gradient and Hessian of the loss at iteration $t$, respectively.

We are interested in computing the gradient of the loss after $T$ steps of gradient descent with respect to the learning rate, $\theta$. This quantity is given by $\frac{\partial d \ell}{\partial \theta} = \langle g^{(T)}, \frac{d w^{(T)}}{d \theta}\rangle $. The second term in this inner product tells us how changes in the learning rate affect the final parameter value after $T$ steps. This quantity can be defined recursively using the total derivative:
\begin{eqnarray*}
    \frac{d w^{(T)}}{d\theta} &=& \frac{\partial w^{(T)}}{\partial w^{(T-1)}} \frac{d w^{(T-1)}}{d\theta} - \frac{\partial w^{(T)}}{\partial \theta} \\
    &=& \left (I - H^{(T-1)} \right)\frac{d w^{(T-1)}}{d\theta} - g^{(T-1)}
\end{eqnarray*}
By expanding the above expression from $t=1$ to $t=T$, we get the following expression for the unrolled gradient:
$$ \frac{d \ell(w^{(T)})}{d\theta} = \left\langle g^{(T)}, -\sum_{i = 0}^{T-1} \left(\prod_{j=i+1}^{T-1} (I - \theta H^{(j)}) \right) g^{(i)}\right\rangle. $$
This expression highlights where the exploding outer-gradient comes from: the recursive definition of $ \frac{d w^{(T)}}{d \theta} $ means that computing it will involve a product of the Hessian at every iteration.

This expression makes intuitive sense if we restrict the number of unrolled steps to one. In this case, the unrolled gradient is the negative inner product between the current and previous gradients: $\frac{d \ell(w^{(T)})}{d\theta} = -\langle g^{(T)}, g^{(T-1)} \rangle$. This means that if the current and previous gradients are correlated (have positive inner product), then updating the learning rate in the direction of the negative unrolled gradient means that we should \textit{increase} the learning rate. This makes sense as if the current and previous gradients are correlated, we expect that we should move faster along this direction.

\newpage
\section{Exploding Gradients on 1D Loss Surfaces} \label{app:toy_lr}

\begin{figure}[h!]
    \centering
    \includegraphics[width=6.5in]{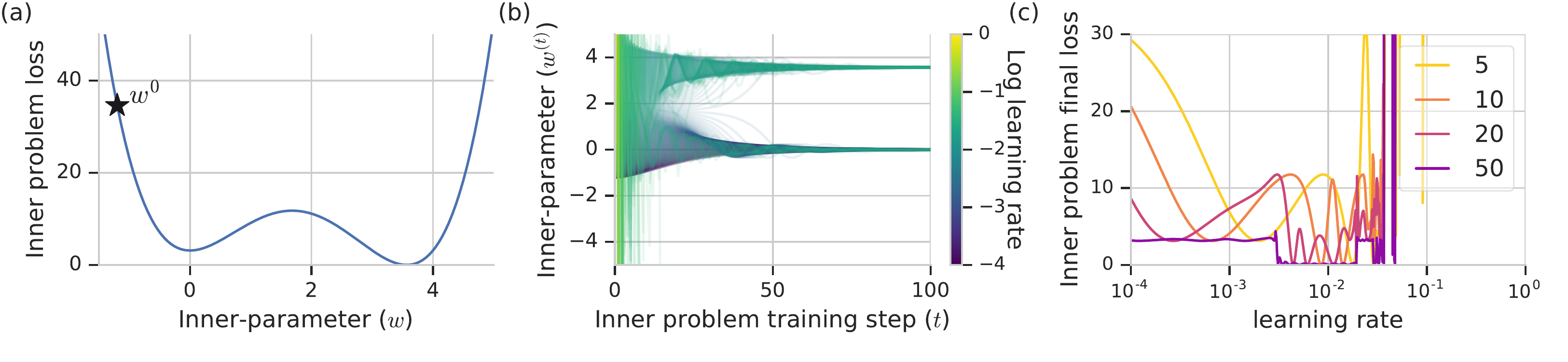}\vspace{-5pt}
    \includegraphics[width=6.5in]{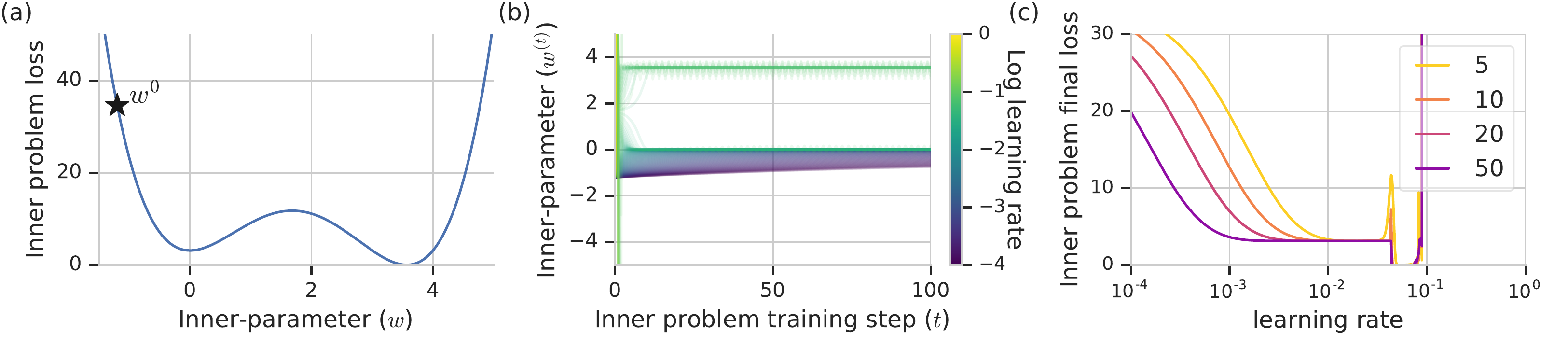}\vspace{-5pt}
    \caption{
    Extension of \ref{fig:2minimum} showing varying learning rate with a fixed momentum (top), and varying learning rate with no momentum (bottom). Outer-problem optimization landscapes can become increasingly pathological with increasing inner-problem step count. 
    \textbf{(a)} A toy 1D inner-problem loss surface with two local minimum. Initial parameter value ($w^{(0)}$) is indicated by the star.
    \textbf{(b)} Final inner-parameter value ($w^{(T)}$) as a function of the number of inner problem training steps $T$. The top row SGD+momentum. Color denotes different values of the optimizer's learning rate parameter. In both settings, low learning rates converge to the first local minimum at $w=0$. Slightly larger learning rates escape this minimum to settle at the global minimum ($w\approx3.5$).
    Even larger values (purples) oscillate before eventually diverging.
    \textbf{(c)} The final loss after some number steps of optimization as a function of the learning rate. 
    Larger values of $T$ result in near discontinuous loss surfaces around the transition points between the two minima.
    \label{fig:2dmin_lr}
    }
\end{figure}

In addition to varying momentum, shown in Figure \ref{fig:2minimum} we also explore the effects of varying learning rate in Figure \ref{fig:2dmin_lr}. 

\newpage
\section{Outer-Training Algorithm}\label{app:algo}
\begin{algorithm}[h!]
  \caption{Outer-training algorithm using the combined gradient estimator.
  \label{alg:in-practice}
  }
\begin{algorithmic}
  \STATE Initialize outer-parameters ($\theta$).
  \WHILE{Outer-training, for each parallel worker}
      \STATE Sample a dataset $\mathcal{D}$, from task distribution $\mathcal{T}$.
      \STATE Initialize the inner loop parameters $w^{\pp{0}}$ randomly.
      \FOR{Each truncation, $t$, in the inner loop}
        \STATE Sample ES perturbation: $e\sim N(0, \sigma^2I)$.
        \STATE Sample a number of steps per truncation, $k$, based on current outer-training iteration.
        \STATE Compute a positive, and negative sequence starting from $w^{\pp{t}}$ by iteratively applying (for $k$ steps), $u(\cdot; \theta+e)$, to $w^{\pp{t}}$
        \STATE Compute a pair of outer-objectives with both a positive, and negative antithetic sample ($L^+$,  $L^-$) using the 2 sequences of $w$ from $t$ to $t+k$ using either the train or validation inner-problem data.
        \STATE Compute a single sample of $g^{rp} = \nabla_{\theta} \frac{1}{2}(L^++L^-)$.
        \STATE Compute a single sample of $g^{es} = \frac{1}{2}(L^+ - L^-)\nabla_{\theta}log(N(e;\theta,\sigma^2I))$
        \STATE Store the sample of $(g_{rp}, g_{es})$ in a buffer until a batch of samples is ready.
        \STATE Assign the current inner-parameter $w$ from one of the two sequences with the inner-parameter value from the end of the truncation ($w^{\pp{t+k}}$).
      \ENDFOR
  \ENDWHILE
  \WHILE{Outer-training}
    \STATE When a batch of gradients is available, compute empirical variance and empirical mean of each weight for each estimator.
     \STATE Use equation \ref{eq:combine} to compute the combined gradient estimate.
    \STATE Update outer-parameters with SGD: $\theta \leftarrow \theta - \alpha g_{combined}$ where $\alpha$ is a learning rate.
  \ENDWHILE
\end{algorithmic}
\end{algorithm}

\section{Architecture details}\label{app:arch_details}
    \subsection{Architecture}
    In a similar vein to diagonal preconditioning optimizers, and existing learned optimizers our architecture operates on each parameter independently.
    Unlike other works, we do not use a recurrent model as we have not found applications where the performance gains are worth the increased computation.
    We instead employ a single hidden layer feed forward MLP with 32 hidden units.
    This MLP takes as input momentum terms at a few different decay values: [0.5, 0.9, 0.99, 0.999, 0.9999]. A similar idea has been explored in \cite{lucas2018aggregated}. The current gradient as well as the current weight value are also used as features (2 additional features).
    By passing in weight values, the optimizer can learn to do arbitrary norm weight decay.
    To emulate learning rate schedules, the current training iteration is fed in transformed via applying a tanh squashing functions at different timescales: $\text{tanh}(t/\eta-1)$ where $\eta$ is the timescale. We use 9 timescales logarithmicly spaced from (3, 300k).
    
    All non-time features are normalized by the second moment with regard to other elements in  the ``batch'' dimension (the other weights of the weight tensor). We choose this over other normalization strategies (e.g. batch norm) to preserve directionality. These activations are then passed the into a hidden layer, 32 unit MLP with ReLU activations.
    Many existing optimizer hyperparameters (such as learning rate) operate on an exponential scale.
    As such, the network produces two outputs, and we combine them in an exponential manner: $\text{exp}(\lambda_{exp} o_1)\lambda_{lin}o_2$ making use of two scaling parameters $\lambda_{\text{exp}}$ and $\lambda_{\text{lin}}$ which are both set to $1e-3$. Without these scaling terms, the default initialization yields steps on the order of size 1 -- far above the step size of any known optimizer and result in highly chaotic regions of $\theta$.
    It is still possible to optimize given our estimator, but training is slow and the solutions found are quite different.
    Code for this optimizer can be found at \url{https://github.com/google-research/google-research/tree/master/task_specific_learned_opt}. 

    \subsection{Inner-problem}
    The optimizer targets a 3 layer convolutional neural network with 3x3 kernels, and 32 units per layer.
    The first 2 layers are stride 2, and the 3rd layer has stride 1. We use ReLU activations and glorot initializations \citep{glorot2010understanding}.
    At the last convolutional layer, an average pool is performed, and a linear projection is applied to get the 10 output classes.
    \subsection{Outer-Training}
    We train using the algorithm described in Appendix~\ref{app:algo} using a linear schedule on the number of unrolling steps from 50 - 10k over the course of 5k outer-training iterations. To add variation in length, we additionally shift this length by a percentage uniformly sampled between (-20\%, 20\%).
    We optimize the outer-parameters, $\theta$, using Adam \citep{kingma2014adam} with a batch size of 128 and with a learning rate of 0.003 for the training outer-objective and 0.0003 for the validation outer-objective, and $\beta_1=0.5$(following existing literature on non-stationary optimization \citep{arjovsky2017wasserstein}).
    While both values of learning rate work for both outer-objectives, we find the validation outer-objective to be \emph{considerably} harder, and training is more stable with the lower learning rate.

    \newpage
    \section{Additional inner loop problem learning curves}
        \label{app:inner_loop_figure}
        
    We plot additional learning curves from both the outer-train task distribution and the outer-validation task distribution. See Figure \ref{fig:unroll_on_problem}. Additionally, we supply 4 additional baselines: RMSProp+Reg+Decay hyper parameter searched for validation and training loss over learning rate, learning rate schedule (both exponential and linear), epsilon, and l1/l2 regularization using 1000 random configurations, and SGDMom+Reg+Decay hyper parameter searched for validation and training loss over learning rate, learning rate schedule, momentum, to use Nesterov momentum, and l1/l2 regularization using 1000 random configurations. The search procedure and search space matches the existing Adam+Reg+Decay procedure described in \ref{sec:learned_opt_performance}.

    \begin{figure}[h!]
        \centering
        \includegraphics[width=6.5in]{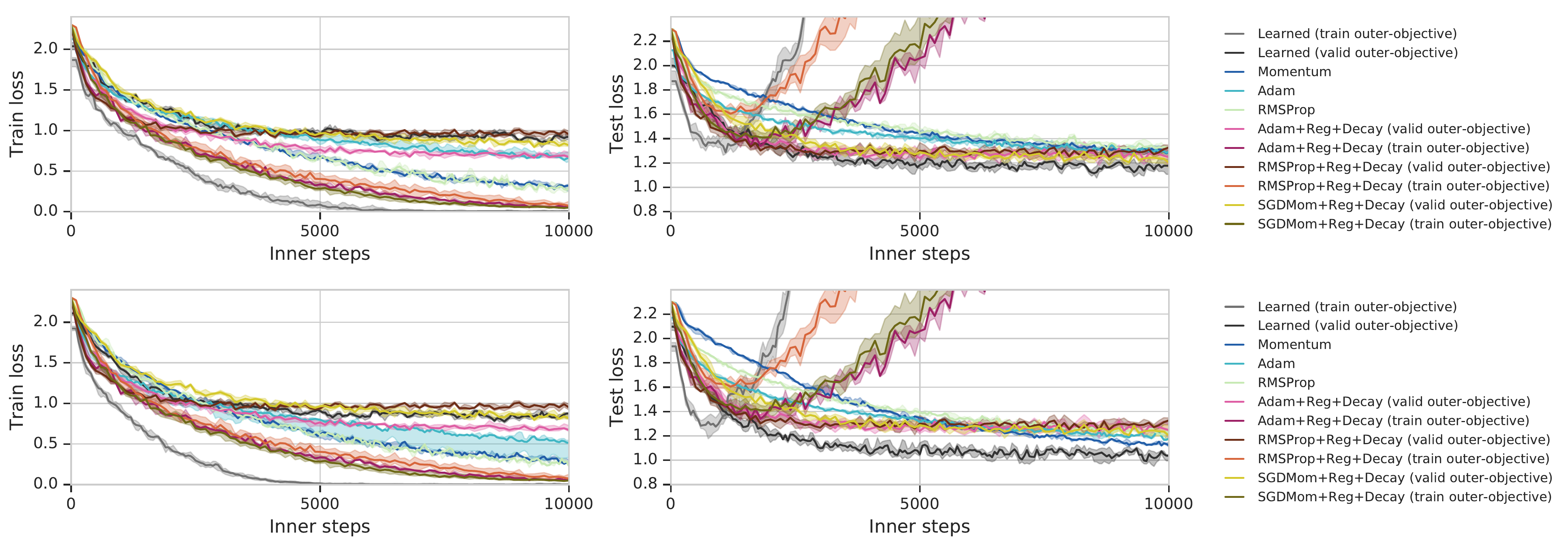}
        \caption{Additional outer-validation problems.
        \label{fig:my_label}
        }
    \end{figure}
    
    \begin{figure}[h!]
        \centering
        \includegraphics[width=6.5in]{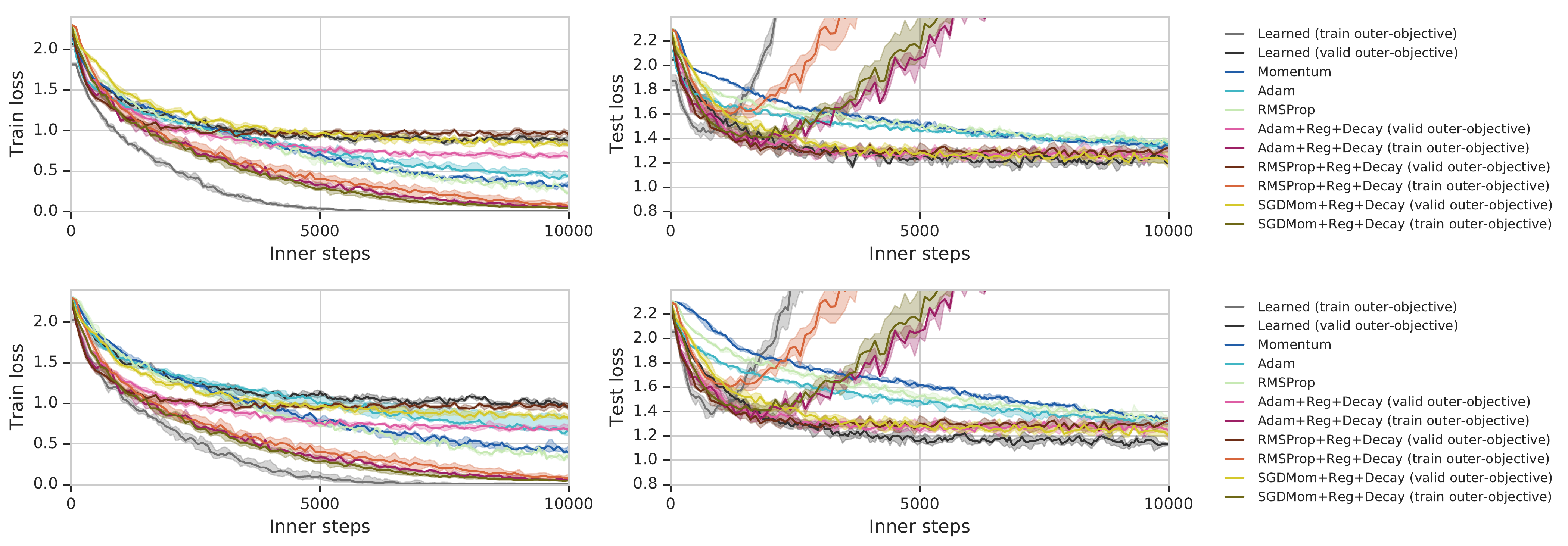}
        \caption{Outer-training problems.
        \label{fig:my_label}
        }
    \end{figure}
    
    \clearpage
    \section{Out of domain generalization} \label{app:outofdist}
    In this work, we focus our attention to learning optimizers over a specific task distribution (3 layer convolutional networks trained on ten class subsets of 32x32 Imagenet).
    In addition to testing on these in domain problems (Appendix \ref{app:inner_loop_figure}), we test our learned optimizer on a variety of out of domain target problems.
    Despite little variation in the outer-training task distribution, our models show promising generalization when transferred to a wide range of different architectures (fully connected, convolutional networks) depths (2 layer to 6 layer) and number of parameters (models roughly 16x more parameters).
    We see these as promising sign that our learned optimizer has a reasonable (but not perfect) inductive bias.
    We leave training with increased variation to encourage better generalization as an area for future work.
   
    \begin{figure}[th!]
        \centering
        \includegraphics[width=5.5in]{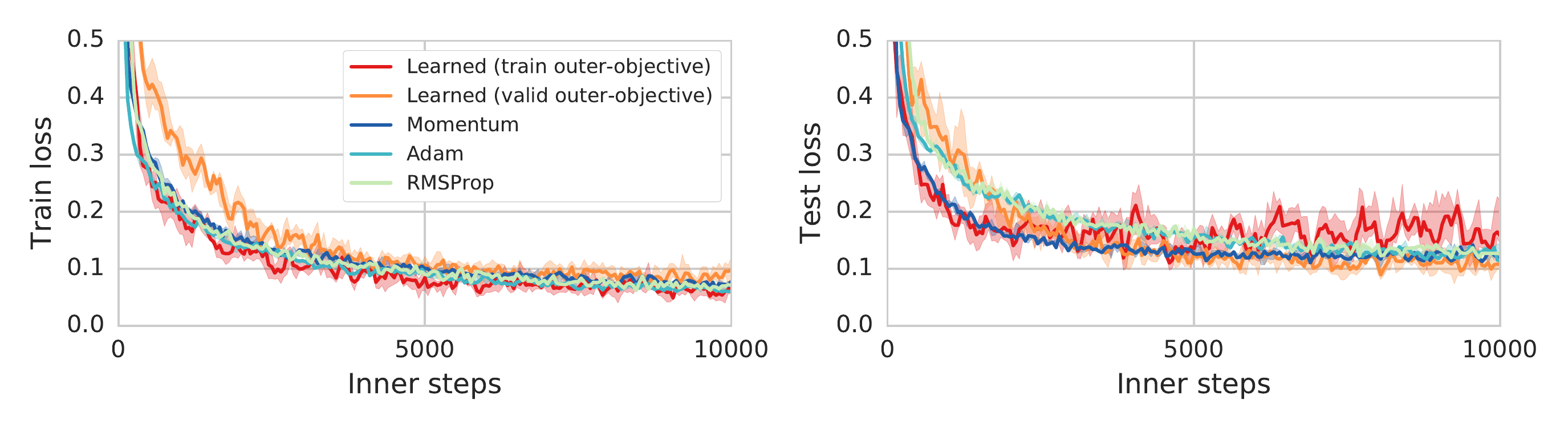}
        \caption{
        Inner problem: 2 hidden layer fully connected network. 32 units per layer with ReLU activations trained on 14x14 MNIST.}
    \end{figure}
    
    \begin{figure}[th!]
        \centering
        \includegraphics[width=5.5in]{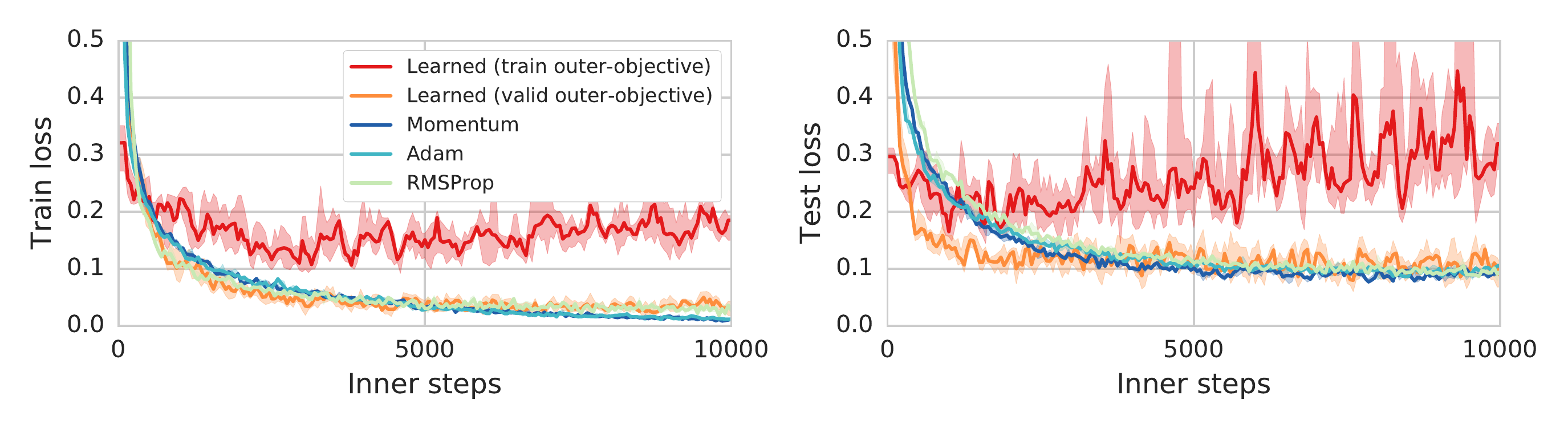}
        \caption{
        Inner problem: 3 hidden layer fully connected network. 128 units per layer with ReLU activations trained on 14x14 MNIST.}
    \end{figure}
    
    \begin{figure}[th!]
        \centering
        \includegraphics[width=5.5in]{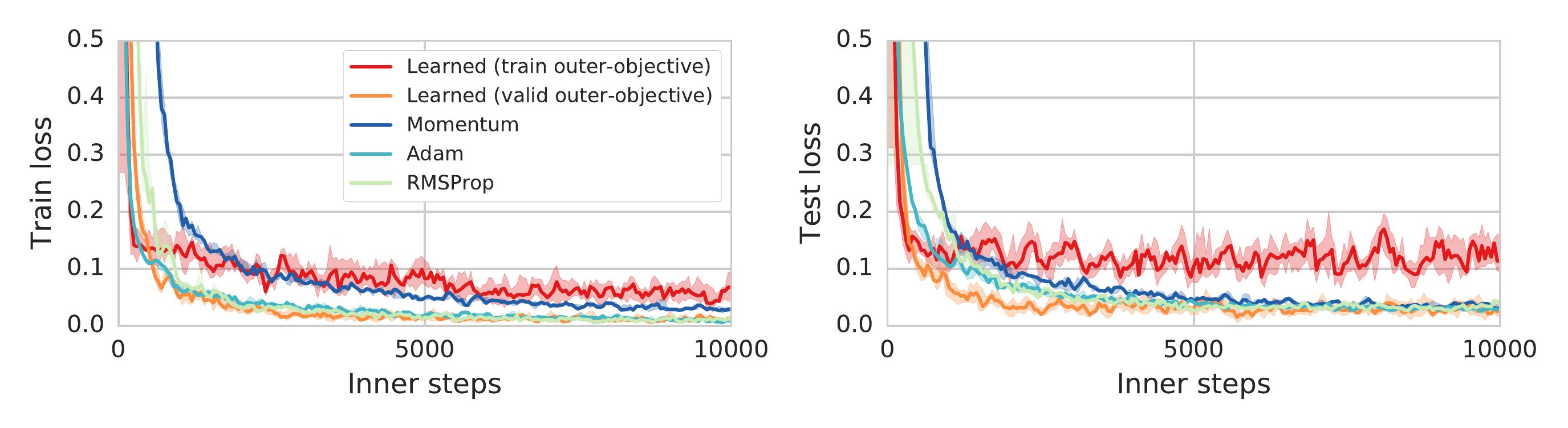}
        \caption{
        Inner problem: 6 convolutional layer network. 32 units per layer, strides: [2,1,2,1,1,1] with ReLU activations on 28x28 MNIST.}
    \end{figure}
    
    \begin{figure}[th!]
        \centering
        \includegraphics[width=5.5in]{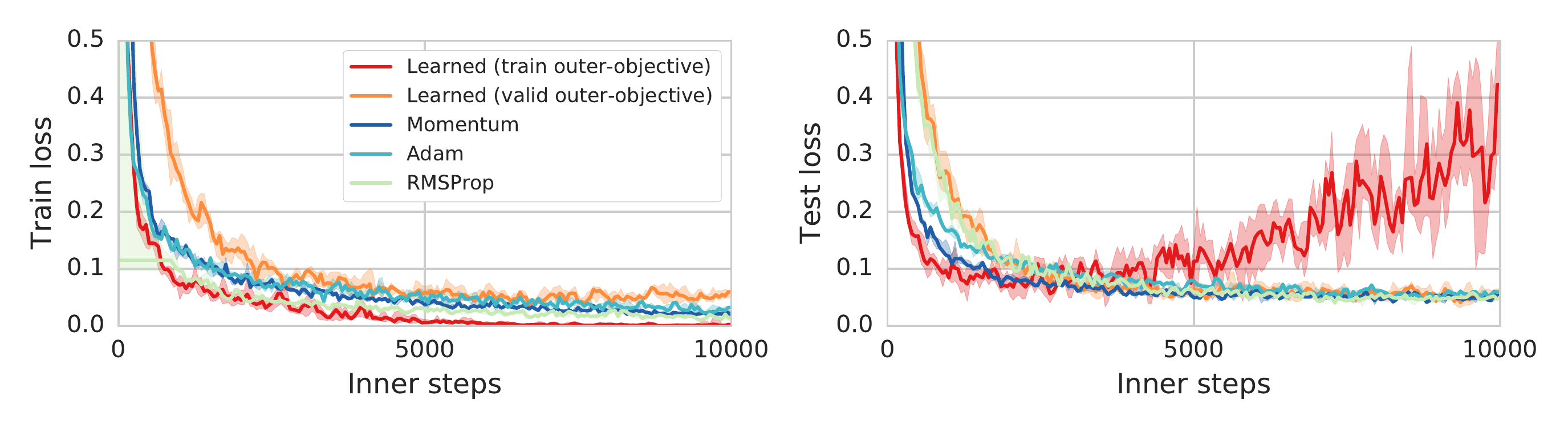}
        \caption{
        Inner problem: 3 convolutional layer network. 32 units per layer, strides: [2,2,1] with ReLU activations on 28x28 MNIST.}
    \end{figure}
    
    \begin{figure}[th!]
        \centering
        \includegraphics[width=5.5in]{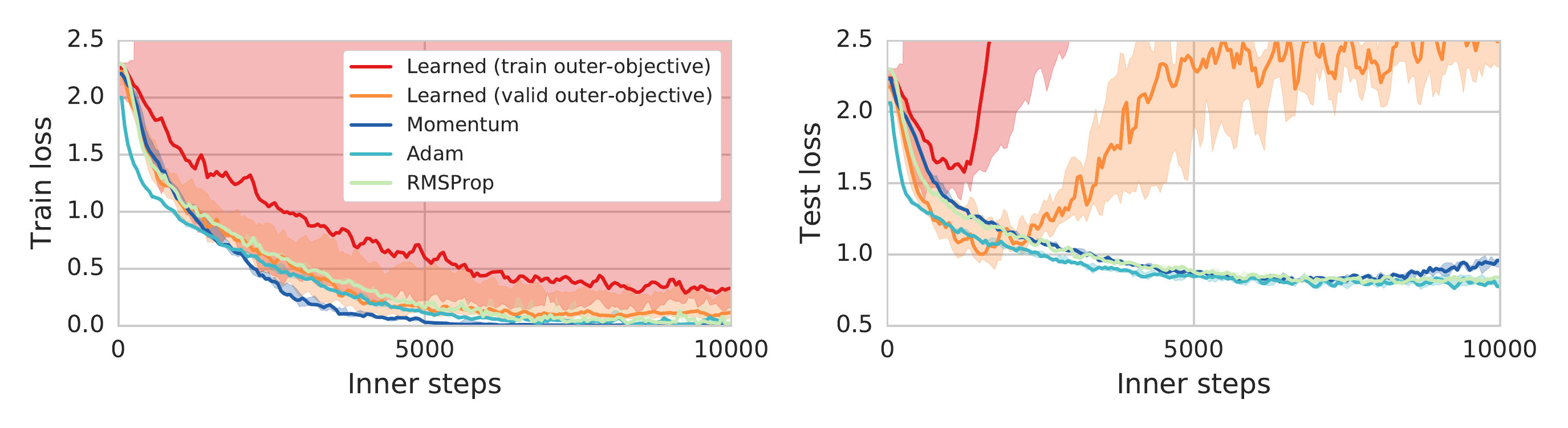}
        \caption{
        Inner problem: 3 convolutional layer network. 128 units per layer, strides: [2,2,1] with ReLU activations trained on a ten-way classification sampled from 32x32 Imagenet (using holdout classes).}
    \end{figure}

    \newpage
    \section{Inner-loop training speed} \label{app:wallclock}
    When training models, often one cares about taking less wall-clock time as compared to loss decrease per weight update.
    Much like existing first order optimizers, the computation performed in our learned optimizer is linear in terms of number of parameters in the model being trained and smaller than the cost of computing gradients.
    The bulk of the computation in our model consists of two batched matrix multiplies of size \textit{features}x32, and 32x2.
    When training models that make use of weight sharing, e.g. RNN or CNN, the computation performed per weight often grows super linearly with parameter count.
    As the learned optimizer methods are scaled up, the additional overhead in performing more complex weight updates will vanish.
    
    For the specific models we test in this paper, we measure the performance of our optimizer on CPU and GPU.
    We re-implement Adam, SGD, and our learned optimizer in TensorFlow (no fused ops) for this comparison.
    Given the small scale of problem we are working at, we implement training in graph in a \textit{tf.while\_loop} to avoid TensorFlow Session overhead.
    We use random input data instead of real data to avoid any data loading confounding.
    On CPU the learned optimizer executes at 80 batches a second where Adam runs at 92 batches a second and SGD at 93 batches per second. The learned optimizer is 16\% slower than both.
    
    On a GPU (Nvidia Titan X) we measure 177 batches per second for the learned and 278 batches per second for Adam, and 358 for sgd. This is or 57\% slower than Adam and 102\% slower than SGD.
    
    Overhead is considerably higher on GPU due to the increased number of ops, and thus kernel executions, sent to the GPU. We expect a fused kernel can dramatically reduce this overhead.
    Despite the slowdown in computation, the performance gains exceed the slowdown, resulting in an optimizer that is still considerably faster when measured in wall-clock time.
    
    For the wall-clock figures presented in this paper we rescale the step vs performance curves by the steps per second instead of directly measuring wall-clock time. When running evaluations, we perform extensive logging which dominates the total compute costs.

    \clearpage
    \section{Ablation learning curves} \label{app:ablation}
    \begin{figure}[h!]
        \centering
        \includegraphics[width=5.5in]{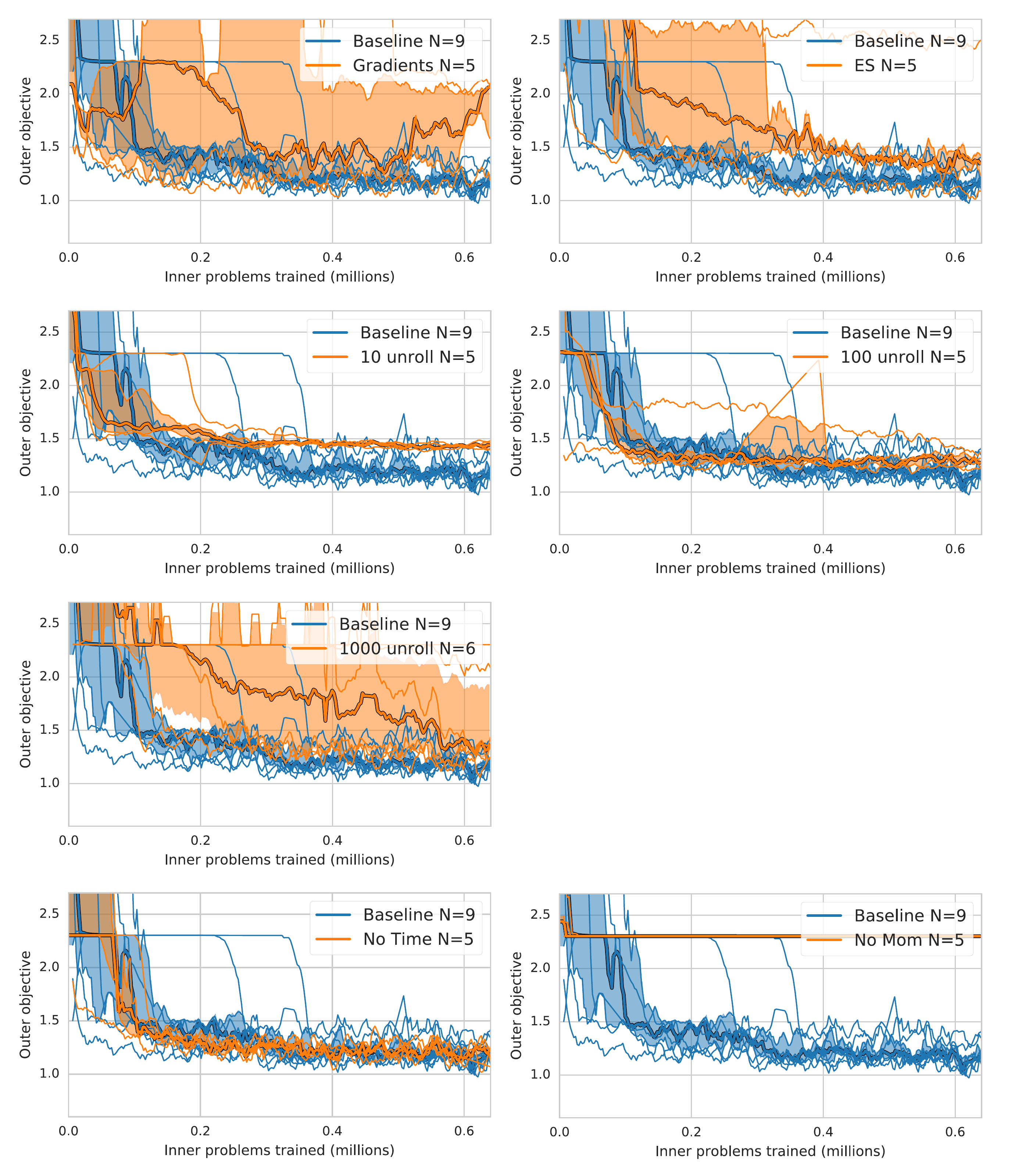}
        \caption{Training curves for ablations described in \sref{sec:ablations}. The thick line bordered in black is the median performance, with the shaded region containing the 25\% and 75\% percentile. Thinner solid lines are individual runs.
        }
    \end{figure}
    \newpage
    \section{Additional Truncation Bias Experiments}  \label{app:trunc_bias}
\begin{figure}[h!]
        \centering
        \includegraphics[width=5.5in]{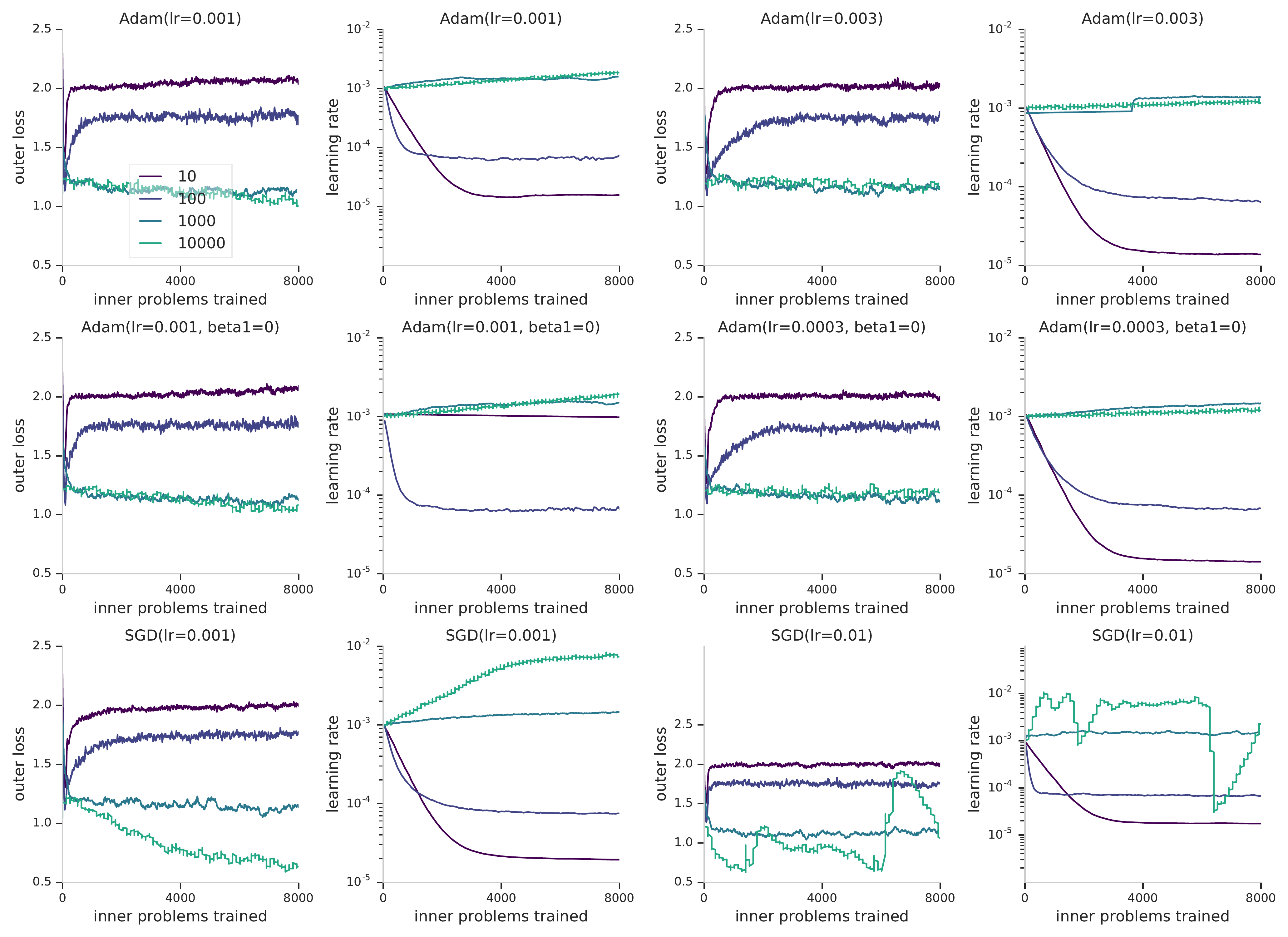}
        \caption{Additional experiments showing truncation bias when attempting to learn Adam hyper-parameters with different outer-optimizers. We show 4 configurations of Adam (2 learning rates, with and without beta1), and 2 configurations of SGD (2 different learning rates) with outer-loss and learning rate plotted for each. In all experiments we see similar, and significant truncation bias when using low amounts of steps per truncation. For some of the higher learning rate experiments (e.g. SGD with lr=0.01), we see diverging loss.\label{fig:different_opt_trunc}
        }
    \end{figure}

    In \sref{sec:increase_bias} we show the effect of truncation bias when learning Adam hyper parameters using the Adam outer-optimizer. When using truncated gradients, the directions passed into this outer-optimizer are not well behaved, and are not even guaranteed to be a conservative vector field. As such, different outer optimizers might behave differently. To test this, we test multiple configurations of outer-optimizer in figure \ref{fig:different_opt_trunc}.
    
    \section{Batch-Normed Base Model}
    In this section, we present experiments targeting a different task family. In particular, we add batch normalization to the convolution layers. We employ the same meta-training procedure and same learned optimizer architecture as used in the rest of this paper and target the validation loss outer-objective.
    
\begin{figure}[h!]
        \centering
        \includegraphics[width=3.5in]{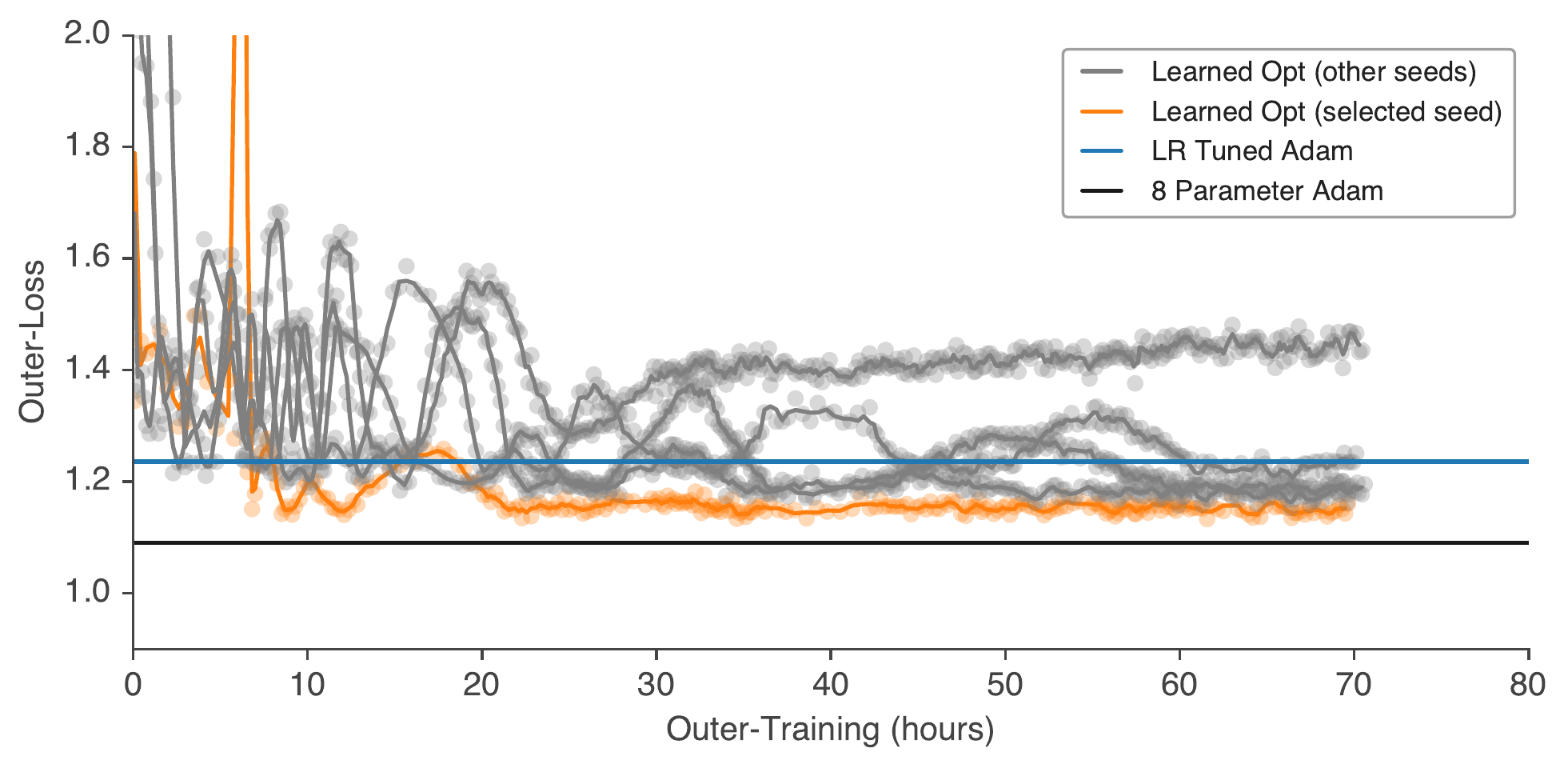}
        \caption{
        Outer-training curves for 5 different random seeds. The learned optimizers outperform LR tuned Adam on 4/5 random seeds but do not yet out perform the 8 parameter Adam baseline. 
        \label{fig:bn_curves}
        }
    \end{figure}

    Meta-training curves can be found in Figure \ref{fig:bn_curves}. We find we outperform learning rate tuned Adam, but do not out perform the 8 parameter tuned Adam baseline. At this point, we are unsure the source of this gap but suspect hyper parameter tuning would improve this result.
    
    \subsection{Longer Unrolls}
\begin{figure}[h!]
        \centering
        \includegraphics[width=5.5in]{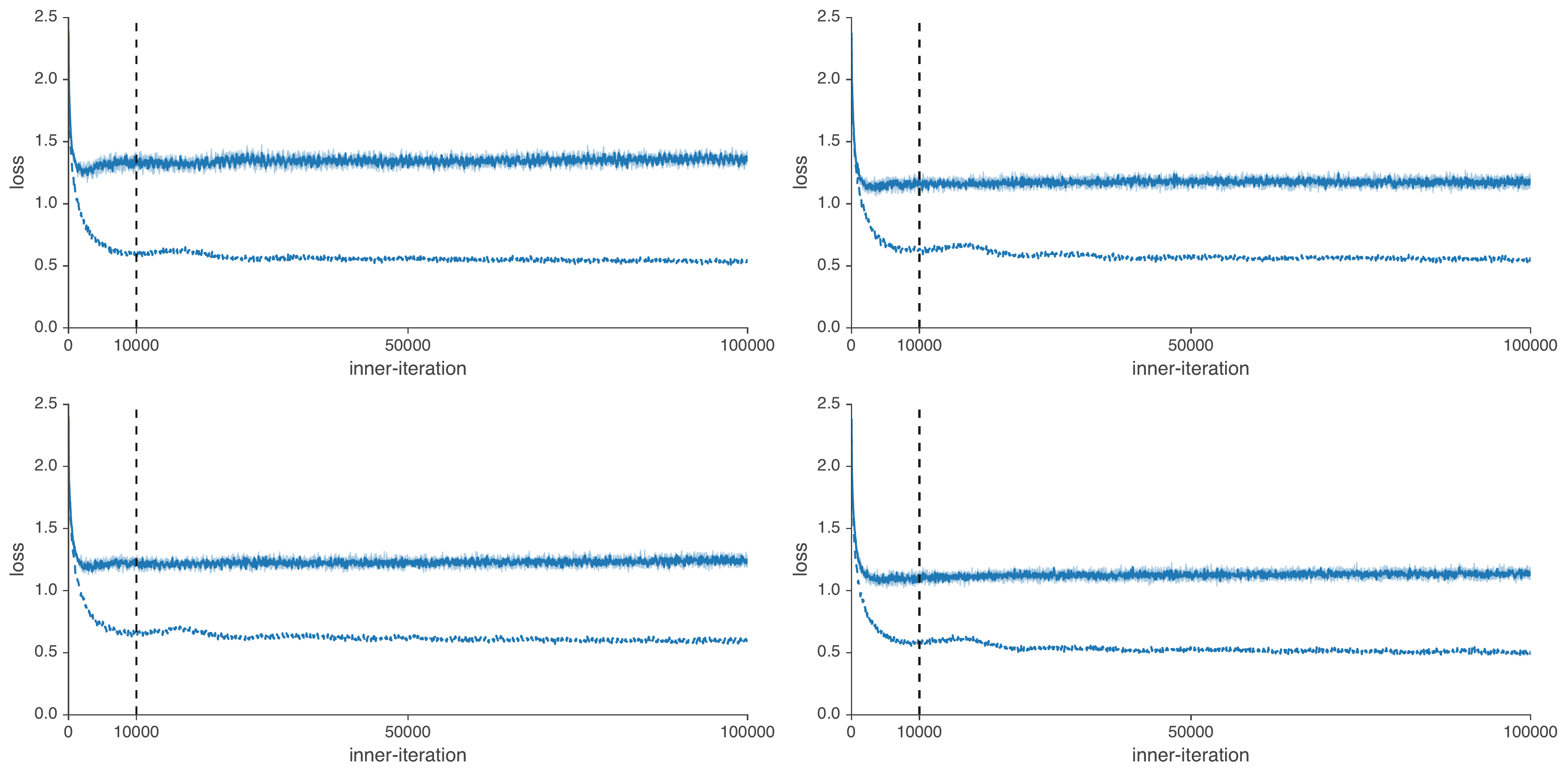}
        \caption{
        Example unrolls on held out outer-tasks when inner-trained for 10x more inner steps. Each panel represents a different held out task. The dashed vertical line denotes the max number of steps seen at outer-training time. The solid line shows inner-test performance, where as the dashed denotes inner-train. We find that our learned optimizers keep a consistent test and train loss even after the 10k iterations used for outer-training.
        \label{fig:longer_unroll}
        }
    \end{figure}

    All of the learned optimizers presented in this paper are outer-trained using 10k inner-iterations. In Figure \ref{fig:longer_unroll} we show an application of a learned optimizer outside of the outer-training regime -- up to 100k inner iterations using the learned optimizer shown in orange in figure \ref{fig:bn_curves}. Unlike hand designed optimizers, our learned optimizer does not completely minimize the training loss. As a result, the test performance remains consistent far outside the outer-training regime.
    
\end{document}